\pdfoutput=1
\documentclass[11pt]{article}

\usepackage[final]{acl}

\usepackage{times}
\usepackage{latexsym}

\usepackage[T1]{fontenc}

\usepackage[utf8]{inputenc}

\usepackage{microtype}

\usepackage{amsfonts}
\usepackage{amssymb}

\usepackage{pifont}
\usepackage{bm}
\usepackage{arydshln}
\usepackage{amsmath}
\usepackage{xcolor}
\usepackage{colortbl}
\usepackage{tcolorbox}
\usepackage{mathtools}
\usepackage{adjustbox}
\usepackage{multirow}
\usepackage{paralist}
\usepackage{makecell}
\usepackage{booktabs}
\usepackage{nicefrac}
\usepackage{enumerate}
\usepackage{bbding} 
\usepackage{wasysym}
\usepackage{inconsolata}
\usepackage{enumitem}
\usepackage{todonotes}
\usepackage{algorithm}
\usepackage{algpseudocode}
\usepackage{booktabs}  
\usepackage{siunitx}   
\usepackage{listings}     
\usepackage{xcolor}       
\usepackage{caption}      
\usepackage{amsmath}      
\usepackage{tcolorbox}
\tcbuselibrary{listings, breakable}
\tcbuselibrary{breakable}
\usepackage{rotating} 
\usepackage{lscape}
\usepackage{graphicx}
\usepackage{subcaption}
\usepackage{booktabs}
\usepackage{makecell}
\usepackage{threeparttable}
\usepackage{wrapfig}

\usepackage{cuted} 
\lstdefinestyle{myjson}{
    basicstyle=\ttfamily\small, 
    breaklines=true,           
    breakatwhitespace=true,    
    showstringspaces=false,
    postbreak=\mbox{\textcolor{red}{$\hookrightarrow$}\space}, 
}
\lstdefinestyle{myprompt}{
    backgroundcolor=\color{black!5},
    basicstyle=\ttfamily\small,
    breaklines=true,
    frame=single,
    framerule=0.5pt,
    framexleftmargin=6pt,
    framexrightmargin=6pt,
    framextopmargin=6pt,
    framexbottommargin=6pt,
    tabsize=2,
    captionpos=b,
    keywordstyle=\color{blue!80!black},
    commentstyle=\color{green!60!black},
    stringstyle=\color{red!80!black},
    showstringspaces=false,
    escapeinside={(*@}{@*)},
}
\definecolor{chong-color}{rgb}{0.858, 0.188, 0.478}

\newcommand\datasetname{\textsc{OPOR-Bench}}
\newcommand\taskname{\textsc{OPOR-Gen}}
\newcommand\evaluationname{\textsc{OPOR-Eval}}

\newtcolorbox{promptbox}[2][]{
	width=\columnwidth,
	colback = gray!8, 
	colframe = gray!8, 
	boxsep=0pt,left=10pt,right=10pt,top=0pt,bottom=0pt,
	fontupper=\linespread{0.9}\selectfont,
	title=#2,#1,
        fontupper=\small}

\definecolor{color1}{RGB}{240,230,140}
\definecolor{color2}{RGB}{197,217,197}
\definecolor{color3}{RGB}{225,179,191}
\definecolor{color4}{RGB}{176,224,230}
\definecolor{where}{RGB}{250,128,114}
\definecolor{strategy}{RGB}{176,224,230}

\definecolor{forestgreen}{RGB}{34, 139, 34}
\definecolor{firebrick}{RGB}{178, 34, 34}

\title{%
    \raisebox{-4pt}{\includegraphics[width=18pt]{"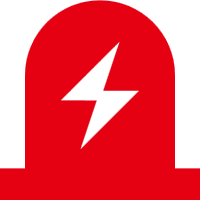"}}%
    OPOR-Bench: Evaluating Large Language Models on Online Public Opinion Report Generation%
}

\author{
  Jinzheng Yu$^{1}$, Yang Xu$^{2}$, Haozhen Li$^{2}$, Junqi Li$^{3}$, Yifan Feng$^{4}$, \\ 
  \textbf{Ligu Zhu}$^{1}$, 
  \textbf{Hao Shen$^{1}$\thanks{Corresponding Authors: shenhao@cuc.edu.cn (H. Shen), leiky\_shi@cuc.edu.cn (L. Shi)}, 
  Lei Shi$^{1}$\footnotemark[1]} \\
  \\ 
  $^{1}$Communication University of China, Beijing, China \\
  $^{2}$Harbin Institute of Technology, Harbin, China \\
  $^{3}$China Academy of Railway Sciences Corporation Limited, Beijing, China \\
  $^{4}$School of Engineering, Santa Clara University, CA, USA \\
  \texttt{\{kimjongyu, zhuligu, shenhao, leiky\_shi\}@cuc.edu.cn} \\
  \texttt{\{yxu, hzli\}@ir.hit.edu.cn} \quad \texttt{lijunqi@rails.cn} \quad \texttt{yfeng3@scu.edu}
  \\[2.5cm] 
}

\begin{document}
\maketitle
\begin{abstract}
Online Public Opinion Reports consolidate news and social media for timely crisis management by governments and enterprises.
While large language models have made automated report generation technically feasible, systematic research in this specific area remains notably absent, particularly lacking formal task definitions and corresponding benchmarks.
To bridge this gap, we define the Automated Online Public Opinion Report Generation (\taskname) task and construct \datasetname, an event-centric dataset covering 463 crisis events with their corresponding news articles, social media posts, and a reference summary.
To evaluate report quality, we propose \evaluationname, a novel agent-based framework that simulates human expert evaluation by analyzing generated reports in context.
Experiments with frontier models demonstrate that our framework achieves high correlation with human judgments.
Our comprehensive task definition, benchmark dataset, and evaluation framework provide a solid foundation for future research in this critical domain.
\end{abstract}
\section{Introduction}\label{sec:intro}
\begin{figure}[t!]
    \centering
    \includegraphics[width=0.48\textwidth]{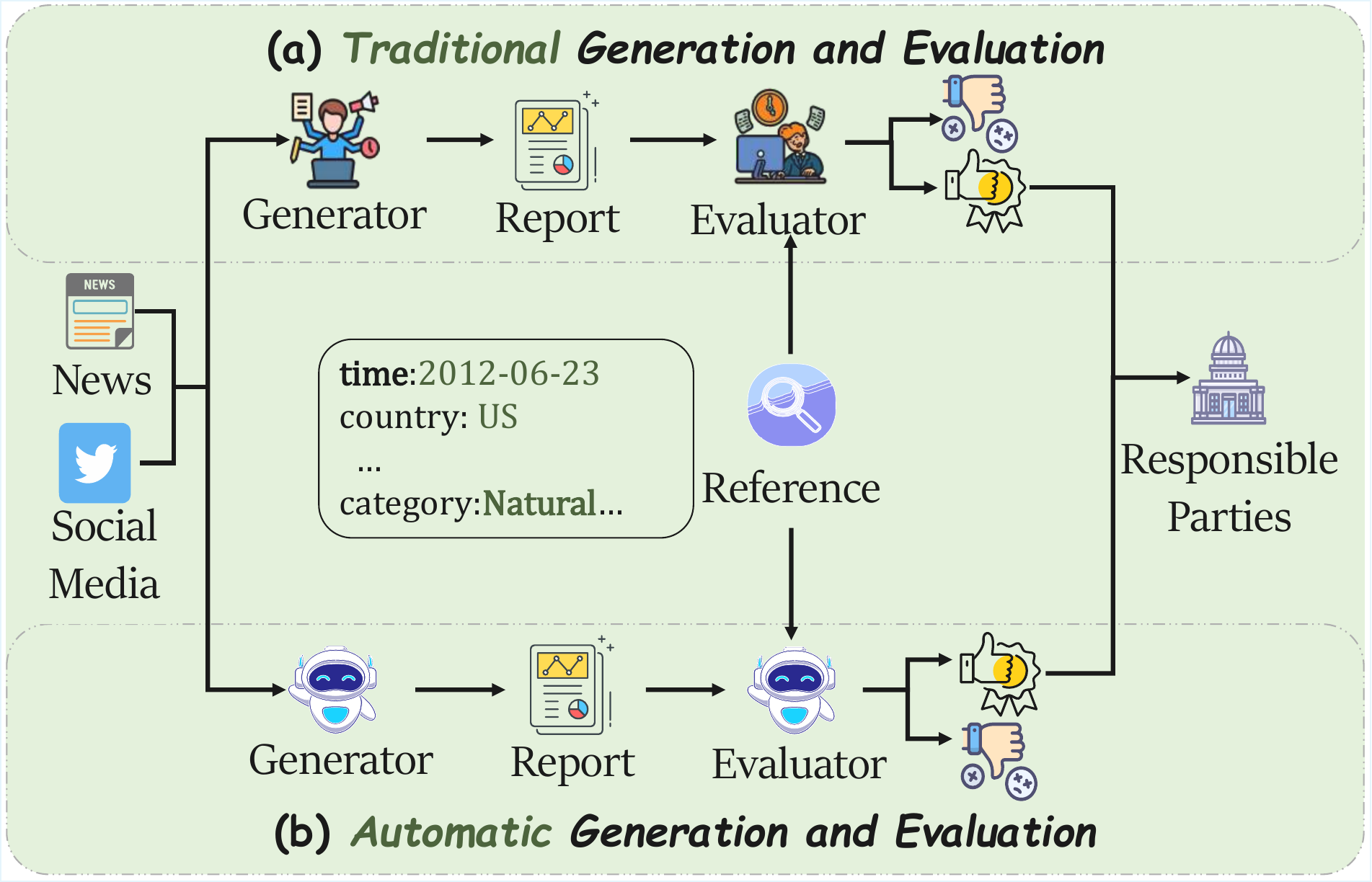} 
    \caption{
(a) Traditional methods require manual information consolidation from diverse sources (e.g., news, social media) and labor-intensive report writing and evaluation. In contrast, our (b) \textbf{automated} approach generates and evaluates reports automatically, significantly accelerating the feedback loop.
}
    \label{fig:compare-task}
\end{figure}

Online Public Opinion Reports are critical tools that consolidate news articles and social media posts about crisis events (e.g., earthquakes, floods) into structured reports, enabling governments or enterprises to respond timely to these rapidly spreading incidents~\citep{Chen2023AboveMS,wang2024esdm,Jin2023HowTF,wang2023cognitive}.  

However, the industry's reliance on manual report generation and evaluation is time-consuming and inefficient, often causing responsible parties to miss optimal response windows and potentially worsen crises.  
The emergence of powerful large language models (LLMs)~\citep{Li2024RiskIO,Li2025GenKPGK,chen2024unlocking} presents a clear opportunity to automate this process. 
Despite this, the problem remains overlooked by the research community, largely because progress is hindered by two fundamental barriers: the lack of a formal task definition for this complex, multi-source generation task, and the absence of a corresponding benchmark dataset.

At its core, the task of Automated Online Public Opinion Report Generation (\taskname) is a sophisticated, domain-specific application of Multi Document Summarization (MDS), as it requires synthesizing vast amounts of information into a coherent document. 
While existing research in MDS has seen significant progress~\citep{Liu2024SumSurveyAA}, \taskname~differs fundamentally from traditional MDS by requiring the generation of a structured, multi-section report from heterogeneous sources (i.e., formal news and informal social media), rather than a single unstructured paragraph from homogeneous texts~\citep{fabbri2019multi}. 
Similarly, while recent LLM-based evaluation methods show promise~\citep{liu2023g,chiang2023can}, they are not tailored to the unique structural and contextual demands of public opinion reports. 


To address these challenges, we define the \taskname~task, which challenges models to synthesize documents from diverse sources (e.g., news and social media) about a single event into a structured report(as shown in Figure~\ref{fig:compare-task}(b)).
We construct \datasetname, the first event-centric dataset designed for the newly defined task.
It comprises 463 crisis events(2012-2025), each containing multi-source documents (news and social media) and a structured reference containing key event information. 

Furthermore, recognizing that evaluating these complex reports is another major challenge, we develop \evaluationname. 
This novel, agent-based framework simulates human expert judgment by leveraging the generated report itself as context to score them on predefined criteria using a 5-point Likert scale.
Given its high agreement with human experts,
this automated approach can replace labor-intensive manual evaluation, significantly accelerating the feedback loop for crisis response~\citep{Cho2024HomeTE}.

This work makes the following key contributions:

\begin{itemize}
\item 
\textbf{A New Task and the First Supporting Benchmark.} 
We define a new task, Automated Online Public Opinion Report Generation (\taskname), and introduce \datasetname, the first event-centric benchmark designed to support it, along with a dedicated annotation tool for quality assurance. 

\item 
\textbf{An Innovative and Reliable Evaluation Framework.} 
We propose \evaluationname, an agent-based framework for evaluating long-form, structured reports, addressing the limitations of traditional metrics. 

\item
\textbf{Comprehensive Baselines and In-depth Analysis.}
We establish strong baselines using frontier models and conduct in-depth analysis of both generation and evaluation. 
Our findings reveal universal challenges in temporal reasoning and systematic evaluation biases, providing concrete directions for future research.

\end{itemize}
\section{Preliminaries: The Structure of an Online Public Opinion Report}

\subsection{Event Title} \label{sec:title}
The Event Title serves as a concise identifier that allows readers to immediately grasp the event's essence. A well-formed title conveys the crisis name, type, and time, facilitating efficient storage and retrieval. 

\subsection{Event Summary} \label{sec:summary}
The Event Summary offers a condensed overview of the crisis for rapid comprehension~\citep{Zhang2021HowTR}. 
Inspired by the classic 5W1H framework~\citep{Wu2025UnfoldingTH, ODay1993OrienteeringIA}, it covers the Crisis Name (What), Location (Where), Time (When), Cause (Why/How), and Involved Parties (Who). 
Crucially, we extend this framework with an Impact component, which highlights the event's consequences to underscore its severity and prompt responsible parties to take action~\citep{Liu2023NetworkPO}.

\subsection{Event Timeline} \label{sec:timeline}
The Event Timeline captures the public opinion lifecycle through three phases: Incubation Period (initial phase with low activity but latent eruption potential), Peak Period (exponential growth with widespread attention and derivative sub-events), and Decline Period (diminishing interest with discussion reverting to affected stakeholders). This lifecycle analysis enables proactive crisis management through early warnings and stage-specific guidance~\citep{Yang2025ForwardingIS,Ren2024SimulationOP}. 

\subsection{Event Focus} \label{sec:focus}
During the Peak Period's volatile environment of polarization and emotional contagion, the Event Focus deconstructs public opinion by analyzing two participant groups—Netizens and Authoritative Institutions.
For each group, the analysis extracts three key insights: (1) Core Topics to reveal their primary concerns; (2) Sentiment Stance to gauge their overall emotional orientation; and (3) Key Viewpoints to highlight their core arguments and stances.

\subsection{Event Suggestions} \label{sec:suggestions}
The Event Suggestions transforms analytical insights into actionable recommendations by synthesizing the event's foundational context (Summary), thematic evolution (Timeline), and divergent viewpoints (Focus). 
These recommendations may include targeted communication strategies, public relations campaigns, or long-term policy adjustments~\citep{Liu2024AnIC,Zhang2021HowTR}.

\section{Task Definition and Dataset}\label{sec:task_and_dataset}
\subsection{Task Definition}
The \taskname~task aims to generate a structured, multi-section report $R_i$ for a given public opinion event $e_i = (X_i, Y_i, Z_i)$. 
The input for each event consists of a set of news articles $X_i = \{x_{i,1}, x_{i,2}, \ldots, x_{i,M}\}$
and a set of social media posts $Y_i = \{y_{i,1}, y_{i,2}, \ldots, y_{i,K}\}$. 

Formally, the task is defined as:
\begin{equation}
R_i = \text{LLM}(P_{\text{gen}}, X_i, Y_i)
\end{equation}
where $P_{\text{gen}}$ represents the generation prompt.

\subsection{\datasetname}
While several crisis-related datasets exist~\citep{olteanu2015expect,imran2016lrec,crisismmd2018icwsm,rsuwaileh2023idrisire,crisisdataset2020icwsm,humaid2020}, they prove inadequate for the \taskname~task due to two critical shortcomings. 
First, they contain only social media posts, which overlooks the formal perspective provided by news articles. 
Second, their effective scale is limited; after thorough quality validation, we find only 95 unique events suitable for our purposes, which is insufficient for a reliable benchmark. 

\begin{figure}[t!]
    \centering
    \includegraphics[width=0.48\textwidth]{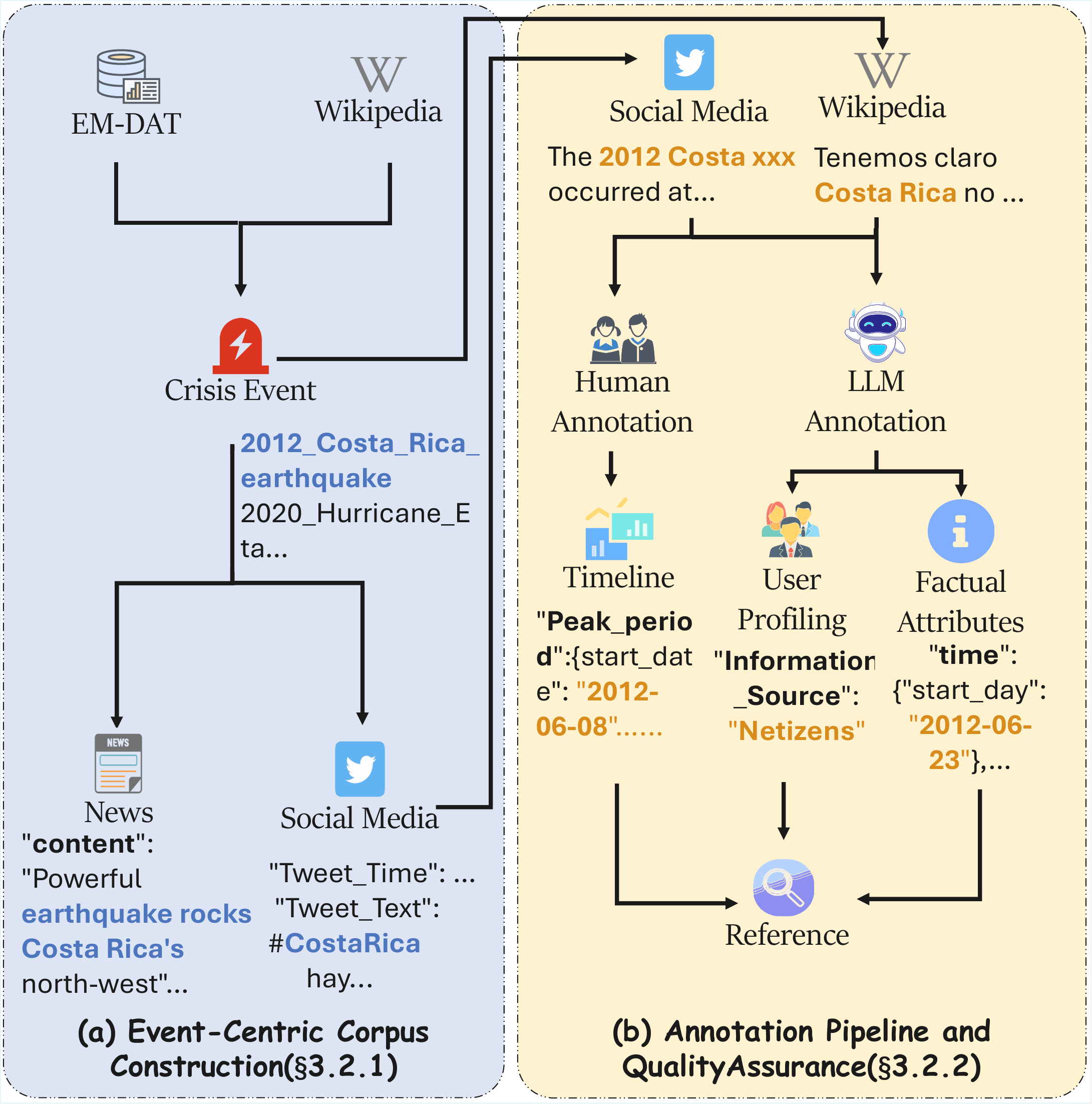} 
    \caption{Overview of \datasetname~construction pipeline. (a) Event-Centric Corpus Construction: Starting from authoritative databases (EM-DAT) and curated lists (Wikipedia), we identify crisis events and collect corresponding multi-source documents—news articles from Wikipedia references and social media posts from X/Twitter API. (b) Dataset Annotation: Three-layered annotation process transforms raw documents into structured data. Human experts annotate timeline phases, while our LLM framework handles factual attribute extraction and social media author classification, ultimately producing a comprehensive reference for each event.}
    \label{fig:dataset-construction}
\end{figure}

\subsubsection{Event-Centric Corpus Construction}
As shown in Figure~\ref{fig:dataset-construction}(a), we begin by collecting crisis events from authoritative sources and then gather documents.

\paragraph{Crisis Event Collection}
The first stage of corpus construction is to identify a large and diverse set of crisis events. 
We gather crisis events (2018-2025) from two sources: the EM-DAT international disaster database~\citep{EMDAT2025,delforge2025dat} for standardized records, and Wikipedia's curated disaster lists for events with high public interest. 
After deduplication against 95 seed events from prior datasets, we obtain 368 new events.
See Appendix~\ref{app:event_sources} for detailed source information.

\paragraph{Document Collection}
For each event, we collect multi-source documents through two parallel streams. 
We gather news articles from the reference lists on each event's official Wikipedia page, while simultaneously acquiring social media posts via the official X (Twitter) API\footnote{\url{https://developer.x.com/en/docs/x-api}} (from one week before to one month after the event). 
The complete collection procedures are  detailed in Appendix~\ref{app:document_collection}.

\subsubsection{Annotation Pipeline and Quality Assurance}
To fulfill the multifaceted requirements of the \taskname~task, we perform three distinct annotation tasks(as shown in Figure~\ref{fig:dataset-construction}(b): (1)Reference Annotation, to extract key factual attributes about each event; (2)Social Media Annotation, to classify the author type of each post; and (3)Timeline Annotation, to pinpoint the start and end dates of each key phase in the public opinion lifecycle. The first two tasks are automated via our protocol-guided LLM framework, while the third, more complex task is handled by human experts to ensure the highest quality.

\paragraph{LLM-based Annotation Framework}
To address the prohibitive expense and time of manual annotation, we develop a protocol-guided LLM framework with three key components: (1)clear label definitions, (2)detailed annotation criteria, and (3)diverse few-shot examples. 
Through preliminary experiments on CrisisLexT26~\citep{olteanu2015expect}, we select gpt-4o-mini for optimal quality-cost balance (see Appendix~\ref{app:llm_selection} for model comparison).

\paragraph{Reference and Social Media Annotation}
In the Reference Annotation task, our framework distills key factual attributes (e.g., location, time, cause) from Wikipedia pages to produce $Z_i$, the structured metadata used for factual evaluation. The subsequent Social Media Annotation task classifies tweet authors into \textit{Netizens} or \textit{Authoritative Institutions}, providing a crucial prerequisite for Event Focus section. 
Detailed protocols for both tasks are available in Appendix~\ref{app:reference_and-social-media_prompt}.

\paragraph{Human Timeline Annotation}
Given the intricate nature of identifying public opinion phases, the timeline annotation is performed entirely by human experts to guarantee annotation quality.
We employ six in-house experts using a dedicated annotation tool\footnote{Our dedicated annotation tool significantly streamlines the process by addressing two primary challenges: (1)providing a unified interface where annotators can view all documents and visualizations simultaneously, and (2)automatically enforcing standardized JSON output to ensure data consistency and eliminate manual formatting errors. Internal tests show this tool reduces annotation time per event by over 50\%.} that visualizes tweet volumes and enforces standardized output. 
See Appendix~\ref{app:timeline_protocol} for detailed procedures.

\subsubsection{Dataset Statistics and Analysis}
\paragraph{Volume and Length Distribution}
As shown in Table~\ref{tab:dataset_stats},  the \datasetname~provides comprehensive multi-source coverage for each crisis event. 
The substantial token count per event (averaging 32K+) highlights the significant information distillation challenge inherent in the \taskname~task.

\begin{table}[ht]
    \centering
    \begin{threeparttable}
        \small 
        \setlength{\tabcolsep}{4pt} 
        \begin{tabular}{lcccc}
            \toprule
             & \textbf{Events} & \textbf{News} & \textbf{Tweets} & \textbf{Reference} \\
            \midrule
            \makecell{Total Num} & 463 & 8,842 & 185,554 & 463 \\
            \makecell{Avg. Num \\  per Event} & - & 19.1 & 400.8 & 1.0 \\
            \makecell{Avg. Token \\ Length\tnote{*}} & 32,531 & 787.2 & 42.5 & 471.0 \\
            \bottomrule
        \end{tabular}
        \caption{Statistics of the \datasetname~ dataset. Token lengths are measured using the cl100k\_base tokenizer from tiktoken.}
        \label{tab:dataset_stats}
    \end{threeparttable}
\end{table}

\begin{figure}[t!]
    \centering
    \includegraphics[width=\linewidth]{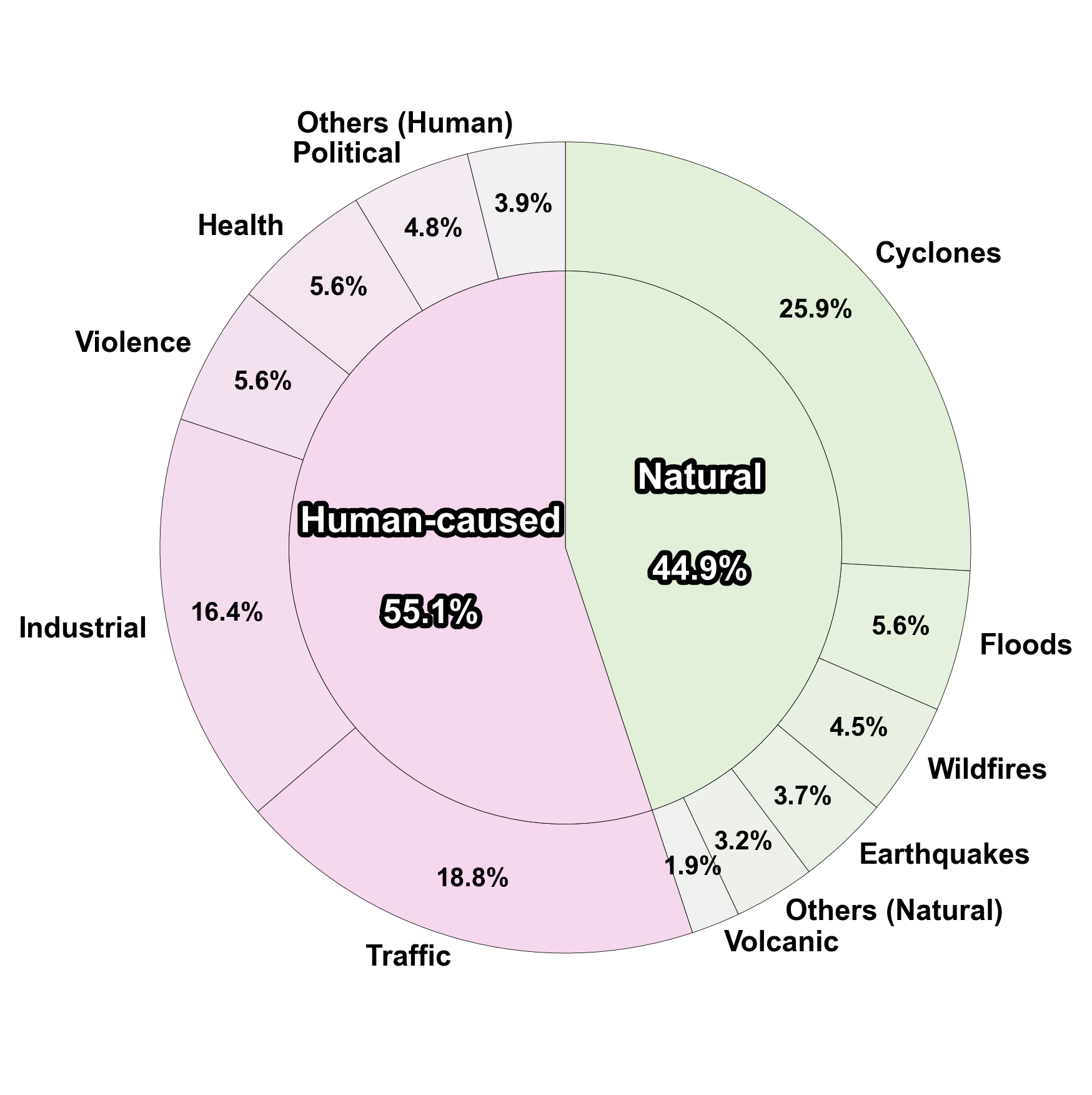} 
    \caption{The distribution of crisis event types in our dataset. The \textbf{inner} ring shows the top-level classification into Natural Disasters (44.9\%) and Human-caused Crises (55.1\%). The \textbf{outer} ring displays the breakdown into more specific sub-categories.}
    \label{fig:event_types}
\end{figure}

\begin{table}[ht!]
    \centering
    \small
    \setlength{\tabcolsep}{3pt}
    \begin{tabular}{
        l
        S[table-format=3.0,tight-spacing=true]
        S[table-format=2.1,tight-spacing=true]
        S[table-format=3.1,tight-spacing=true]
    }
        \toprule
        \textbf{Event Sub-type} & \textbf{\# Events} & \textbf{\# Avg. News} & \textbf{\# Avg. Tweets} \\
        \midrule
        \multicolumn{4}{l}{\textit{Natural Disasters}} \\
        Cyclones & 120 & 21.3 & 400.9 \\
        Floods & 26 & 14.0 & 551.3 \\
        Wildfires & 21 & 16.9 & 768.2 \\
        Earthquakes & 17 & 20.7 & 761.2 \\
        Volcanic & 9 & 24.2 & 210.3 \\
        Others (Natural) & 15 & 25.2 & 854.5 \\
        \multicolumn{4}{l}{\textit{Human-caused Crises}} \\
        Traffic & 87 & 16.5 & 274.8 \\
        Industrial & 76 & 15.2 & 341.5 \\
        Health & 26 & 22.0 & 455.0 \\
        Violence & 26 & 22.5 & 268.3 \\
        Political & 22 & 24.6 & 114.0 \\
        Others (Human) & 18 & 17.5 & 442.3 \\
        \bottomrule
    \end{tabular}
    \caption{Detailed statistics for each event sub-type, including the number of events and the average number of associated news articles and social media posts.}
    \label{tab:type_richness}
\end{table}

\paragraph{Type and Category Distribution} 
Figure~\ref{fig:event_types} shows our dataset's balanced coverage of natural disasters (44.9\%) and human-caused crises (55.1\%). Table~\ref{tab:type_richness} reveals distinct media coverage patterns across event types—"Political" events generate more news coverage while "Wildfires" trigger higher social media activity—underscoring the necessity of multi-source integration.

\begin{table}[htbp]
    \centering
    \small 
    \begin{tabular}{%
        l                               
        S[table-format=3.0]             
        S[table-format=3.1]             
    }
        \toprule
        \textbf{Continent} & \textbf{\# Events} & \textbf{Percentage} \\
        \midrule
        Asia          & 185 & 40.0 \\
        North America & 136 & 29.4 \\
        Europe        &  58 & 12.5 \\
        Africa        &  48 & 10.4 \\
        Oceania       &  21 &  4.5 \\
        South America &  15 &  3.2 \\
        \bottomrule
    \end{tabular}
    \caption{Geographic distribution of the \datasetname~ dataset across the six populated continents.}
    \label{tab:continent_distribution}
\end{table}

\paragraph{Geographic Distribution} 
Table~\ref{tab:continent_distribution} shows our dataset spans 108 countries across six continents, ensuring geographic diversity and mitigating potential regional bias.

\section{\evaluationname} \label{sec:evaluationname}
\subsection{Overview}
In our framework, the evaluator role (shown in Figure~\ref{fig:compare-task}(b)) is realized through an agent-based architecture that manages three specialized tools.
Given a generated report $R_i$, its associated social media posts $Y_i$, and reference $Z_i$, the evaluation task produces a 15-dimensional score vector $S_i = (s_{i,1}, s_{i,2}, \ldots, s_{i,15})$, where each score $s_{i,j}$ is rated on a 5-point Likert scale:

\begin{equation}
S_i = \text{LLM}(P_{\text{eval}}, R_i, Y_i, Z_i)
\end{equation}

where $P_{\text{eval}}$ is the evaluation prompt containing detailed scoring criteria. 

\begin{figure}[t]
    \centering
    \includegraphics[width=\linewidth]{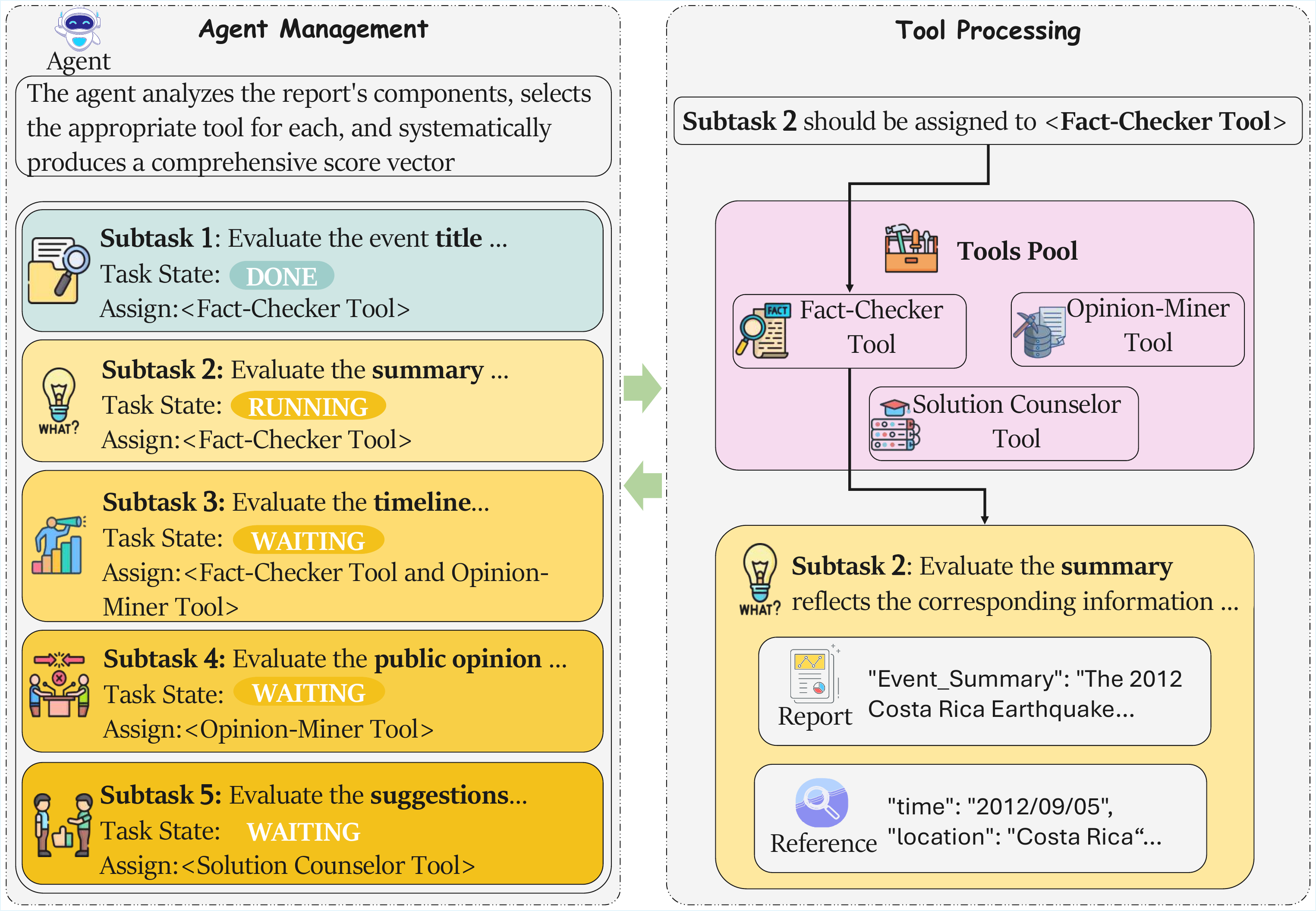} 
    \caption{The \evaluationname~architecture: An evaluation agent manages three specialized tools (Fact-Checker, Opinion-Miner, Solution-Counselor) through structured task assignment.}
    \label{fig:evaluation_framework}
\end{figure}

Our 15 dimensions are established through iterative expert consensus.
As shown in Figure~\ref{fig:evaluation_framework}, the \evaluationname~agent (\S\ref{sec:Agent}) employs three specialized tools: (1) \textbf{Fact-Checker} (\S\ref{sec:Fact-Checker}) for verifying accuracy against references, (2) \textbf{Opinion-Miner} (\S\ref{sec:Opinion-Miner}) for evaluating public opinion coverage, and (3) \textbf{Solution-Counselor} (\S\ref{sec:Solution Counselor}) for evaluating recommendations. 
This transforms evaluation from black-box judgment into a traceable analytical process~\citep{hu2024hiagent,wang2024s3}. 
See Appendix~\ref{app:evaluation_criteria} for detailed scoring guidelines.

\subsection{Architecture}

\subsubsection{\evaluationname~Agent} \label{sec:Agent}
The \evaluationname~Agent manages the evaluation process by analyzing report components, selecting appropriate tools, and producing the comprehensive score vector. Following a reasoning-acting protocol~\citep{Yao2022ReActSR}, the agent explicitly externalizes reasoning before each action 
(see Appendix~\ref{app:prompt-agent} for implementation details).

This process is formalized as:
\begin{equation}
\label{eq:agent_func}
S_i = \text{Agent}(R_i, Y_i, Z_i) = S_{i, \text{fact}} \oplus S_{i, \text{opin}} \oplus S_{i, \text{sol}}
\end{equation}
where $\oplus$ denotes vector concatenation and the components are produced by each specialized tool. 

\subsubsection{Fact-Checker tool} \label{sec:Fact-Checker}
The \textbf{Fact-Checker Tool} ($T_{\text{fact}}$) verifies the factual accuracy of the report's \textit{Title}, \textit{Summary}, and the \textit{Date Accuracy} of the \textit{Event Timeline} by comparing them against the reference data ($Z_i$). 
This tool evaluates the model's \textbf{Factual Consistency}—its ability to stay faithful to the provided source material.
\begin{equation}
\label{eq:tool_fact}
\begin{split}
T_{\text{fact}}(R_{i, \text{title}}, R_{i, \text{summary}}, R_{i, \text{timeline}}, Z_i) \\
= S_{i, \text{fact}}
= (s_{i,1}, s_{i,2}, \dots, s_{i,7}) 
\end{split}
\end{equation}
See Appendix~\ref{app:fact_checker_criteria} for scoring guidelines.

\subsubsection{Opinion-Miner Tool} \label{sec:Opinion-Miner}
The \textbf{Opinion-Miner Tool} ($T_{\text{opin}}$) evaluates the \textit{Sub-Events} of the \textit{Event Timeline} and the entire \textit{Event Focus} section by comparing them against the source social media posts ($Y_i$).
It measures the model's \textbf{Multi-source Synthesis}—extracting key insights from large volumes of noisy, unstructured text.
\begin{equation}
\label{eq:tool_opin}
\begin{split}
T_{\text{opin}}(R_{i, \text{timeline}}, R_{i, \text{focus}}, Y_i) \\
= S_{i, \text{opin}}
= (s_{i,8}, s_{i,9}, s_{i,10}, s_{i,11}) 
\end{split}
\end{equation}
Detailed scoring guidelines are in Appendix~\ref{app:opinion_miner_criteria}.

\subsubsection{Solution Counselor Tool} \label{sec:Solution Counselor}
The \textbf{Solution Counselor Tool} ($T_{\text{sol}}$) leverages the LLM's internal knowledge and reasoning to evaluate the report's \textit{Event Suggestions} based on criteria such as their feasibility, relevance, and innovation.
This directly tests the model's \textbf{Practical Reasoning}—generating novel, actionable solutions from parametric knowledge.
\begin{equation}
\label{eq:tool_sol}
T_{\text{sol}}(R_{i, \text{suggestions}}) \\
= S_{i, \text{sol}}
= (s_{i,12}, s_{i,13}, s_{i,14}, s_{i,15}) 
\end{equation}

The detailed scoring guidelines are available in Appendix~\ref{app:solution_counselor_criteria}.

\section{Experiments} \label{sec:experiment}
\subsection{Experimental Setup}
\subsubsection{Models and Strategies}
We evaluate five frontier LLMs with 128K+ context windows: GPT-4o~\citep{openai2024gpt4o}, DeepSeek-R1~\citep{deepseek2025r1}, DeepSeek-V3~\citep{deepseek2024v3}, Gemini 2.5 Pro~\citep{google2024gemini2_5}, and Llama-3.3-70B~\citep{meta2024llama3}.

\begin{figure}[t]
    \centering
    \includegraphics[width=\linewidth]{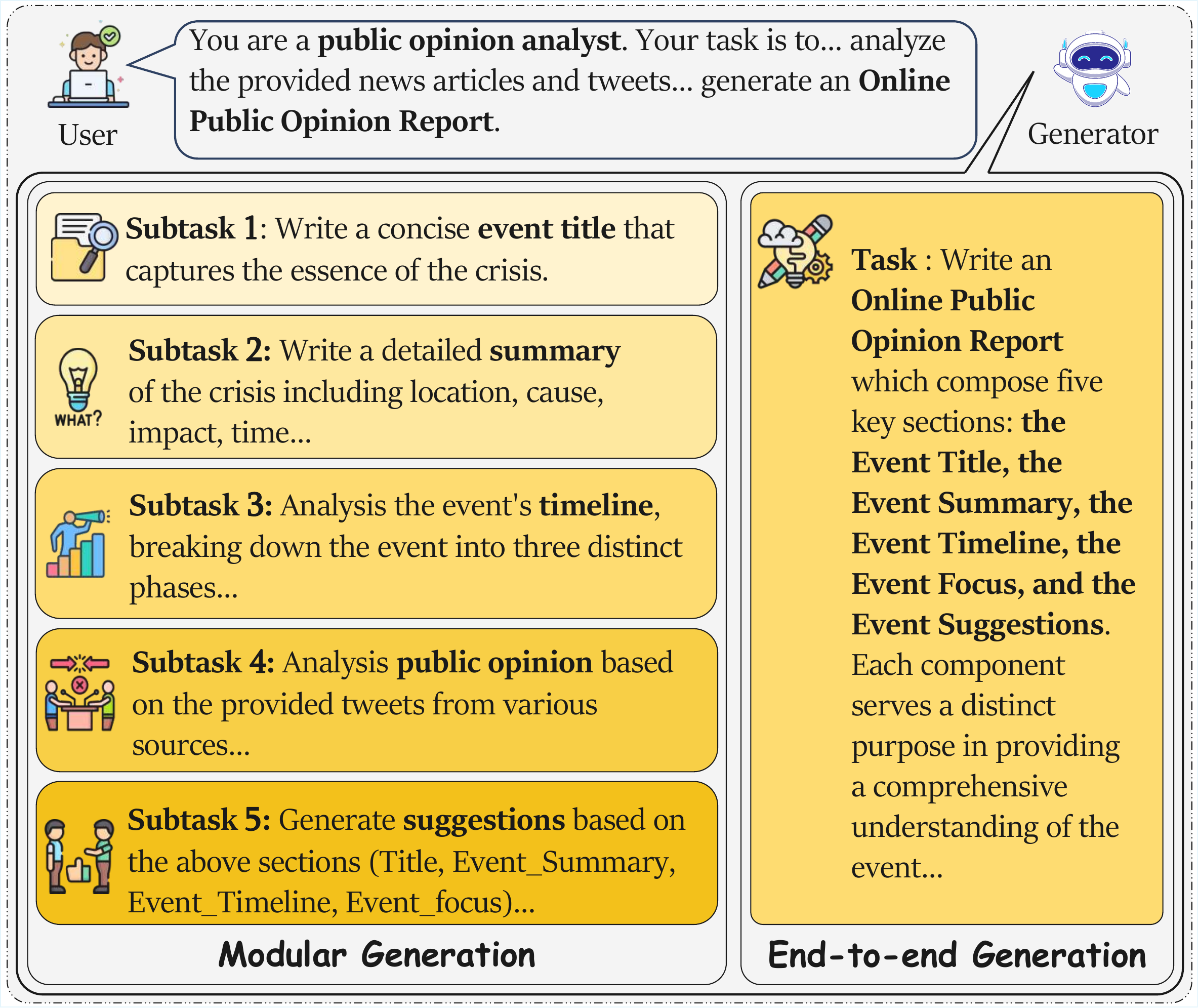}
    \caption{Comparison of two \taskname~strategies. \textbf{Left}: Modular generation decomposes the task into five sequential subtasks (title, summary, timeline, focus, and suggestions), with each component generated independently. \textbf{Right}: End-to-end generation produces all five report components simultaneously in a single pass, maintaining global coherence throughout the document.}
    \label{fig:generation_strategies}
\end{figure}

For each model, we employ two distinct generation strategies(Figure~\ref{fig:generation_strategies}): \textbf{modular} generation creating sections independently then assembling them~\citep{bai2024longwriter}, and \textbf{end-to-end} generation producing complete reports in a single pass.

\begin{table}[t]
   \centering
   \small 
   \begin{tabular}{lc}
       \toprule
       \textbf{Dimension} & \textbf{ICC3} \\
       \midrule
       \textit{Event Title} & 0.843 \\
       \textit{Event Summary} & \\
       \quad Event Nature & 0.868 \\
       \quad Time \& Loc. & 0.860 \\
       \quad Involved Parties & 0.856 \\
       \quad Causes & 0.879 \\
       \quad Impact & 0.887 \\
       \textit{Event Timeline} & \\
       \quad Date Acc. & 0.839 \\
       \quad Sub Events & 0.793 \\
       \textit{Event Focus} & \\
       \quad Contro. Topic & 0.894 \\
       \quad Repr. Stmt. & 0.877 \\
       \quad Emo. Anal. & 0.893 \\
       \textit{Event Suggestions} & \\
       \quad Rel. & 0.625 \\
       \quad Feas. & 0.621 \\
       \quad Emo. Guide. & 0.640 \\
       \quad Innov. & 0.676 \\
       \bottomrule
   \end{tabular}
   \caption{The ICC scores confirm high inter-rater reliability among human experts.}
   \label{tab:icc_results}
\end{table}

\begin{table*}[t]
    \centering
    \small
    \begin{tabular}{lccccccc}
        \toprule
        \multirow{2}{*}{\textbf{Dimension}} & \multicolumn{2}{c}{\textbf{Spearman's $\rho$}} & \multicolumn{2}{c}{\textbf{Kendall's $\tau$}} & \multicolumn{2}{c}{\textbf{MAE}} \\
        \cmidrule(lr){2-3} \cmidrule(lr){4-5} \cmidrule(lr){6-7}
        & \textbf{DeepSeek-V3} & \textbf{GPT-4o} & \textbf{DeepSeek-V3} & \textbf{GPT-4o} & \textbf{DeepSeek-V3} & \textbf{GPT-4o} \\
        \midrule
        \textit{Event Title} & 0.55 & 0.72 & 0.48 & 0.63 & 0.64 & 0.54 \\
        \textit{Event Summary} & & & & & & \\
        \quad Event Nature & 0.57 & 0.75 & 0.49 & 0.66 & 0.73 & 0.52 \\
        \quad Time \& Loc. & 0.64 & 0.78 & 0.54 & 0.68 & 0.70 & 0.52 \\
        \quad Involved Parties & 0.45 & 0.83 & 0.37 & 0.73 & 0.96 & 0.45 \\
        \quad Causes & 0.56 & 0.84 & 0.46 & 0.73 & 0.85 & 0.50 \\
        \quad Impact & 0.41 & 0.85 & 0.33 & 0.75 & 0.91 & 0.63 \\
        \textit{Event Timeline} & & & & & & \\
        \quad Date Acc. & 0.18 & 0.80 & 0.14 & 0.70 & 1.36 & 0.45 \\
        \quad Sub Events & 0.19 & 0.75 & 0.16 & 0.64 & 1.02 & 0.58 \\
        \textit{Event Focus} & & & & & & \\
        \quad Contro. Topic & 0.07 & 0.81 & 0.06 & 0.70 & 1.24 & 0.54 \\
        \quad Repr. Stmt. & 0.15 & 0.77 & 0.12 & 0.66 & 1.20 & 0.52 \\
        \quad Emo. Anal. & 0.11 & 0.82 & 0.09 & 0.71 & 1.12 & 0.51 \\
        \textit{Event Suggestions} & & & & & & \\
        \quad Rel. & 0.11 & 0.47 & 0.09 & 0.41 & 0.64 & 0.47 \\
        \quad Feas. & - & 0.22 & - & 0.19 & 0.57 & 0.56 \\
        \quad Emo. Guide. & 0.11 & 0.50 & 0.10 & 0.43 & 0.60 & 0.59 \\
        \quad Innov. & 0.04 & 0.52 & 0.03 & 0.45 & 0.71 & 0.59 \\
        \bottomrule
    \end{tabular}
    \caption{GPT-4o demonstrates superior human-agent alignment over DeepSeek-V3.}
    \label{tab:human_agent_agreement}
\end{table*}

\begin{table*}[t]
    \centering
    \small
    \setlength{\tabcolsep}{4pt} 
    \begin{tabular}{lll|ccccccc}
        \toprule      
        
        \textbf{Evaluator} & \textbf{\parbox{1.3cm}{\centering Gen\\Strategy}} & \textbf{\parbox{2.2cm}{\centering Model}} & \textbf{\parbox{1.2cm}{\centering Event\\Title}} & \textbf{\parbox{1.3cm}{\centering Event\\Summary}}  & \textbf{\parbox{1.3cm}{\centering Event\\Timeline}} & \textbf{\parbox{1.2cm}{\centering Event\\Focus}} & \textbf{\parbox{1.7cm}{\centering Event\\Suggestions}} & \textbf{\parbox{1.0cm}{\centering Avg.\\Score}} \\
        \midrule
        \multirow{10}{*}{DeepSeek-V3} 
         & \multirow{5}{*}{End-to-end} 
         & DeepSeek-R1 & \textbf{4.48} & 4.25 & 3.20 & 4.07 & 4.16 & 4.03 \\
         & & DeepSeek-V3 & 4.37 & 4.19 & 2.99 & 4.02 & 4.12 & 3.94 \\
         & & Gemini 2.5 Pro & 4.37 & \textbf{4.27} &\textbf{3.40} & \textbf{4.27} & \textbf{4.20} & \textbf{4.10} \\
         & & GPT-4o & 4.29 & 4.10 & 3.06 & 4.03 & 4.05 & 3.91 \\
         & & Llama-3.3-70B & 4.15 & 3.95 & 2.67 & 3.89 & 3.95 & 3.72 \\
         \cmidrule{2-9}
         & \multirow{5}{*}{Modular} 
         & DeepSeek-R1 & 4.34 & \textbf{4.18} & \textbf{3.56} & 4.25 & \textbf{4.22} & \textbf{4.11} \\
         & & DeepSeek-V3 & 4.28 & 4.14 & 3.46 & 4.23 & 4.17 & 4.06 \\
         & & Gemini 2.5 Pro & \textbf{4.38} & 4.15 & 3.52 & \textbf{4.29} & 4.17 & 4.10 \\
         & & GPT-4o & 4.12 & 3.99 & 3.40 & 4.21 & 4.12 & 3.97 \\
         & & Llama-3.3-70B & 4.15 & 3.86 & 3.20 & 4.19 & 4.02 & 3.88 \\
         \midrule
        \multirow{10}{*}{GPT-4o} 
         & \multirow{5}{*}{End-to-end} 
         & DeepSeek-R1 & \textbf{4.50} & 3.65 & 2.56 & 3.75 & 3.99 & 3.69 \\
         & & DeepSeek-V3 & 4.40 & 3.57 & 2.52 & 3.75 & 3.99 & 3.65 \\
         & & Gemini 2.5 Pro & 4.47 & \textbf{3.73} & \textbf{2.71} & \textbf{3.85} & \textbf{4.00} & \textbf{3.75} \\
         & & GPT-4o & 4.33 & 3.52 & 2.56 & 3.77 & 3.95 & 3.63 \\
         & & Llama-3.3-70B & 4.27 & 3.42 & 2.24 & 3.68 & 3.92 & 3.51 \\
         \cmidrule{2-9}
         & \multirow{5}{*}{Modular} 
         & DeepSeek-R1 & 4.16 & 3.55 & 2.67 & \textbf{3.85} & \textbf{4.01} & 3.65 \\
         & & DeepSeek-V3 & 4.14 & 3.54 & 2.63 & 3.78 & 3.97 & 3.61 \\
         & & Gemini 2.5 Pro & \textbf{4.26} & \textbf{3.57} & \textbf{2.70} & 3.84 & 3.98 & \textbf{3.67} \\
         & & GPT-4o & 4.04 & 3.42 & 2.68 & 3.84 & 3.97 & 3.59 \\
         & & Llama-3.3-70B & 4.13 & 3.27 & 2.56 & 3.74 & 3.94 & 3.53 \\
        \bottomrule
    \end{tabular}
    \caption{\textbf{Overall} performance comparison of five LLMs using two  generation strategies (end-to-end and modular), evaluated by two  distinct LLM evaluators (DeepSeek-V3 and GPT-4o). Within each experimental block, the highest score for each evaluation dimension is  highlighted in \textbf{bold}.}

    \label{tab:performance_summary_0}
\end{table*}

\subsubsection{Implementation Details}
We use temperature 0.7 for generation and 0.3 for evaluation tasks. All other hyperparameters follow model defaults. See Appendix~\ref{app:prompt_templates} for prompt templates.

\subsubsection{Evaluation Setup}
We implement \evaluationname~with GPT-4o and DeepSeek-V3 to evaluate the generated reports across 15 dimensions following identical protocols (Appendix~\ref{app:evaluation_criteria}). 
Additionally, we conduct human evaluation on a subset of these reports to validate the overall effectiveness.

\subsection{Evaluation Framework Validation} 
\subsubsection{Human Evaluation Protocol}
To validate \evaluationname~, we conduct a two-phase human evaluation with three experts who also design the scoring criteria. Using a dedicated annotation tool (Figure~\ref{fig:human_evaluation_tool}) ensuring blind evaluation, experts independently score each report. The protocol includes calibration and formal evaluation phases, detailed in Appendix~\ref{app:human_eval_phases}.

\subsubsection{Agreement Analysis and Results}
In this section, we analyze the results of our human evaluation to answer two key questions: (1) How reliable are our human experts? and (2) How strong is the agreement between our human experts and the \evaluationname~agents?

\paragraph{Answer 1: Our human experts demonstrate high inter-rater reliability.} 
Table~\ref{tab:icc_results} shows high inter-rater reliability among human experts. 
Following established guidelines where Intraclass Correlation Coefficient (ICC) values are categorized as poor (<0.50), moderate (0.50-0.75), good (0.75-0.90), and excellent (>0.90)~\citep{Koo2016AGO}, most dimensions achieve good to excellent agreement with ICC>0.75. 
The relatively lower agreement on Event Suggestions (ICC=0.64, moderate) reflects the inherent subjectivity in evaluating recommendation quality.


\paragraph{Answer 2: The \evaluationname~framework achieves strong human-agent alignment with GPT-4o.}
Following common guidelines, correlation is categorized as weak (<0.50), moderate (0.50-0.70), or strong (>0.70)~\citep{Cohen1969StatisticalPA}. For MAE, lower values indicate better alignment~\citep{Willmott2005AdvantagesOT}. Our human-agent agreement analysis (Table~\ref{tab:human_agent_agreement}) reveals that the GPT-4o achieves a strong overall alignment ($\rho$=0.69, MAE=0.53). In contrast, DeepSeek-V3 shows moderate performance, with notably poor correlation to subjective Opinion Mining tasks ($\rho$=0.13).  This confirms that while \evaluationname~is effective with a strong model, human oversight remains valuable for these subjective dimensions.


\subsection{General Results}


\paragraph{Overall Performance} 
Given GPT-4o's strong alignment with human judgment, we report the average of its scores across both generation strategies to determine each model's final performance. As shown in Table~\ref{tab:performance_summary_0}, \textbf{Gemini 2.5 Pro} leads with an average score of 3.71, followed by DeepSeek-R1 (3.67), DeepSeek-V3 (3.63), GPT-4o (3.61), and Llama-3.3-70B (3.52). The narrow 5.12\% gap between the best and worst models indicates they all have a solid baseline capability for the \taskname~task.

\paragraph{Comparison of Generation Strategies}
As shown in Table~\ref{tab:performance_summary_0}, while the end-to-end generation strategy slightly outperforms the modular approach on average (GPT-4o: 3.65 vs. 3.61), our results reveal a clear trade-off: end-to-end excels at high-level synthesis (Title, Summary), while the modular approach is superior for detailed, multi-perspective analysis (Timeline, Focus). 
This key finding suggests that the optimal strategy is task-dependent, pointing towards hybrid approaches as a promising direction for future research.

\section{Analysis and Discussion}







\subsection{Task Complexity Analysis}

\subsubsection{Temporal Reasoning is a Universal Challenge for LLMs.}
Analysis of GPT-4o evaluation results (Table~\ref{tab:performance_summary_0}) reveals a notable weakness across all models in the \textbf{Event Timeline} dimension. 
This weakness is rooted in a dramatic failure on the \textbf{Date Accuracy} sub-dimension (average: 1.25; see details in Appendix~\ref{app:main-results}, Table~\ref{tab:performance_summary_1}).
This failure is consistent even for the top-performing model (Gemini 2.5 Pro: 1.28) and stems from the task's demand for complex temporal reasoning—identifying inflection points in data trends rather than simply extracting dates from documents.

\begin{figure}[t]
    \centering
    \includegraphics[width=0.93\linewidth]{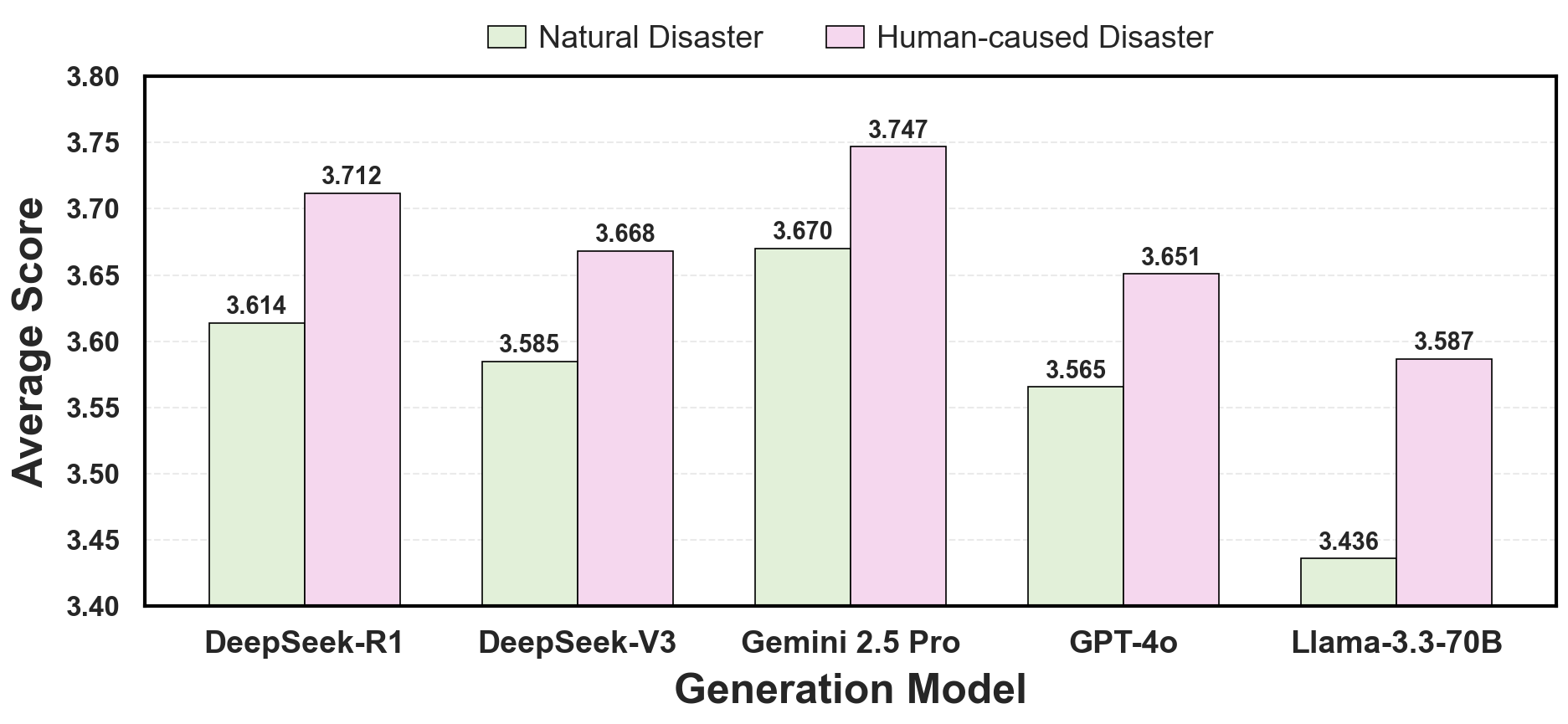}
    \caption{Consistent performance gap between human-caused (higher) and natural disasters (lower) across all models.}
    \label{fig:rq-big}
\end{figure}

\begin{figure}[!htbp]
    \centering
    \includegraphics[width= \linewidth]{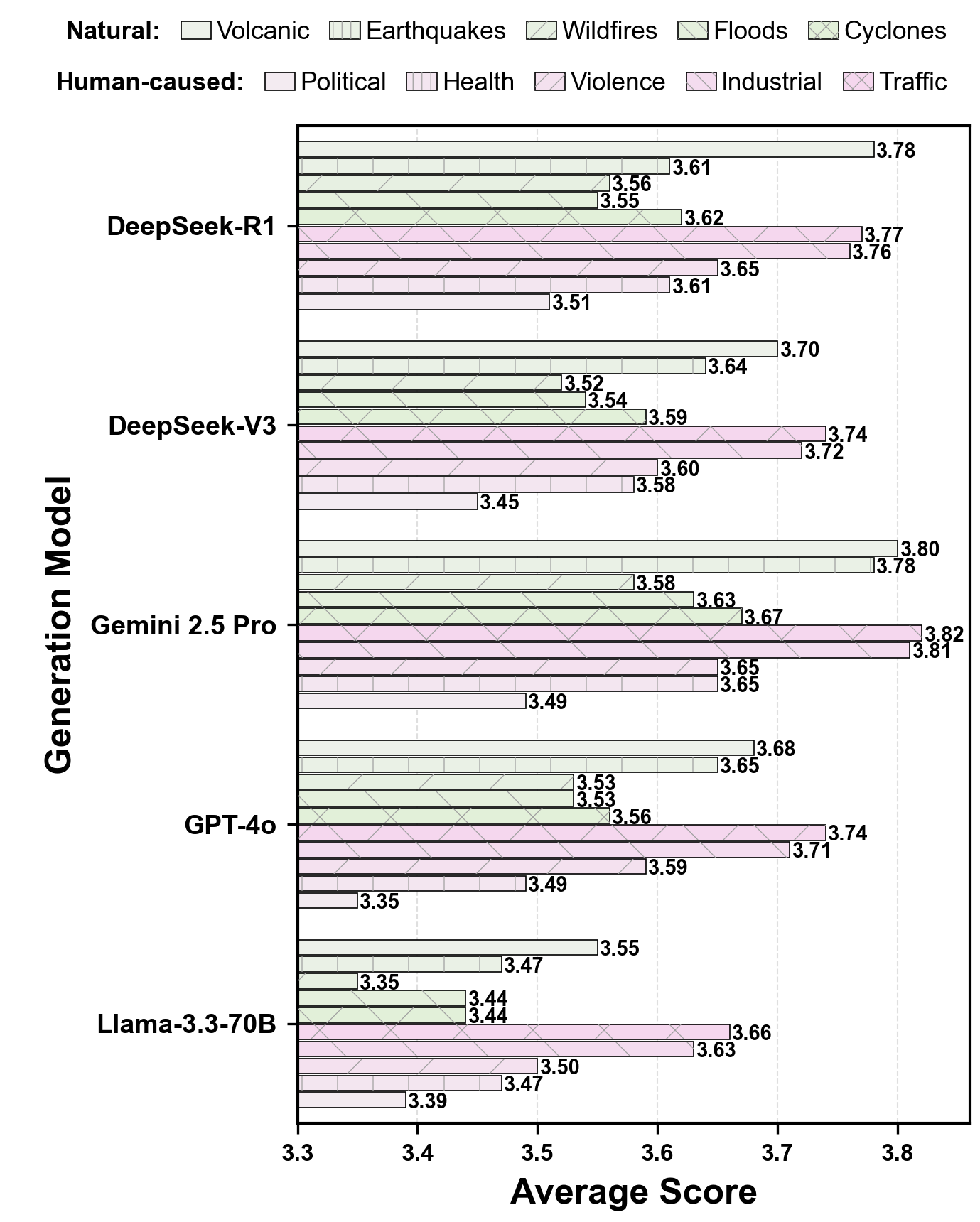}
    \caption{Performance correlates with information structure: structured events (Industrial, Traffic) score highest while diffuse events (Wildfires, Floods) score lowest.}
    \label{fig:rq-small}
\end{figure}

\subsubsection{Difficulty Stems from Information Structure, Not Thematic Content.}
A consistent pattern across all models (Figure~\ref{fig:rq-big}) suggests this relationship: performance is significantly higher for Human-caused Disasters than for Natural Disasters.
Further examination at the sub-category level (Figure~\ref{fig:rq-small}) reveals a clearer distinction: models excel on events with well-defined information structures, like "Industrial" and "Traffic" accidents, but struggle with events characterized by diffuse information, such as "Wildfires" and "Floods." 
We hypothesize that this disparity stems from the fact that human-caused disasters typically feature clear causal chains and structured data (e.g., official investigation reports) that are easily processed by LLMs. In contrast, natural disasters generate fragmented information from diverse sources with ambiguous temporal boundaries, posing a fundamental challenge to fact extraction and timeline segmentation.

\begin{table}[t]
    \centering
    \small
    \setlength{\tabcolsep}{3.5pt} 
    \begin{tabular}{l S[table-format=1.2] S[table-format=1.2] S[table-format=1.2]}
        \toprule
        \textbf{Model} & \textbf{\parbox{1.4cm}{\centering Factual\\Consistency}} & \textbf{\parbox{1.6cm}{\centering Multi-source\\Synthesis}} & \textbf{\parbox{1.6cm}{\centering Practical\\Reasoning}} \\
        \midrule
        DeepSeek-R1     & \textbf{3.06} & 3.89 & \textbf{4.00} \\
        DeepSeek-V3     & 3.02 & 3.84 & 3.98 \\
        Gemini 2.5 Pro  & 3.10 & \textbf{3.99} & 3.99 \\
        GPT-4o          & 2.99 & 3.87 & 3.96 \\
        Llama-3.3-70B   & 2.91 & 3.66 & 3.93 \\ 
        \bottomrule
    \end{tabular}
    \caption{Average Model Performance Across Three Evaluation Categories}
    \label{tab:crisis_type_performance}
\end{table}
\begin{table}[t]
    \centering
    \small
    \setlength{\tabcolsep}{2pt} 
    \begin{tabular}{l S[table-format=-1.4] S[table-format=-1.4]} 
        \toprule
        \textbf{Score Category} & {\textbf{\parbox{2cm}{\centering News Count\\Correlation (r)}}} & {\textbf{\parbox{2cm}{\centering Tweet Count\\Correlation (r)}}} \\
        \midrule
        Title       & -0.0342  & {-} \\
        Summary     & -0.1373  & {-} \\
        Timeline    & {-}  &  0.2134 \\
        Focus       & {-}  & -0.4411 \\
        \bottomrule
    \end{tabular}
    \caption{Correlation Between News/Tweet Count and Evaluation Scores}
    \label{tab:news-tweet_count}
\end{table}

\subsection{Generator Performance Analysis} 
\subsubsection{LLMs Performance Is Remarkably Consistent Across All Three Evaluation Categories.} 
Our evaluation framework evaluates three distinct capabilities: Factual Consistency (via Fact-Checker Tool), Multi-source Synthesis (via Opinion-Miner Tool), and Practical Reasoning (via Solution Counselor Tool). 
As shown in Table~\ref{tab:crisis_type_performance}, model performance rankings remain remarkably stable across these three capabilities. Pearson correlations between all category pairs exceed 0.84 (p<0.001), indicating that models' capabilities are comprehensive. This suggests current LLMs possess balanced strengths across factual verification, information integration, and strategic reasoning tasks.

\begin{figure}[t]
    \centering
    \includegraphics[width=\linewidth]{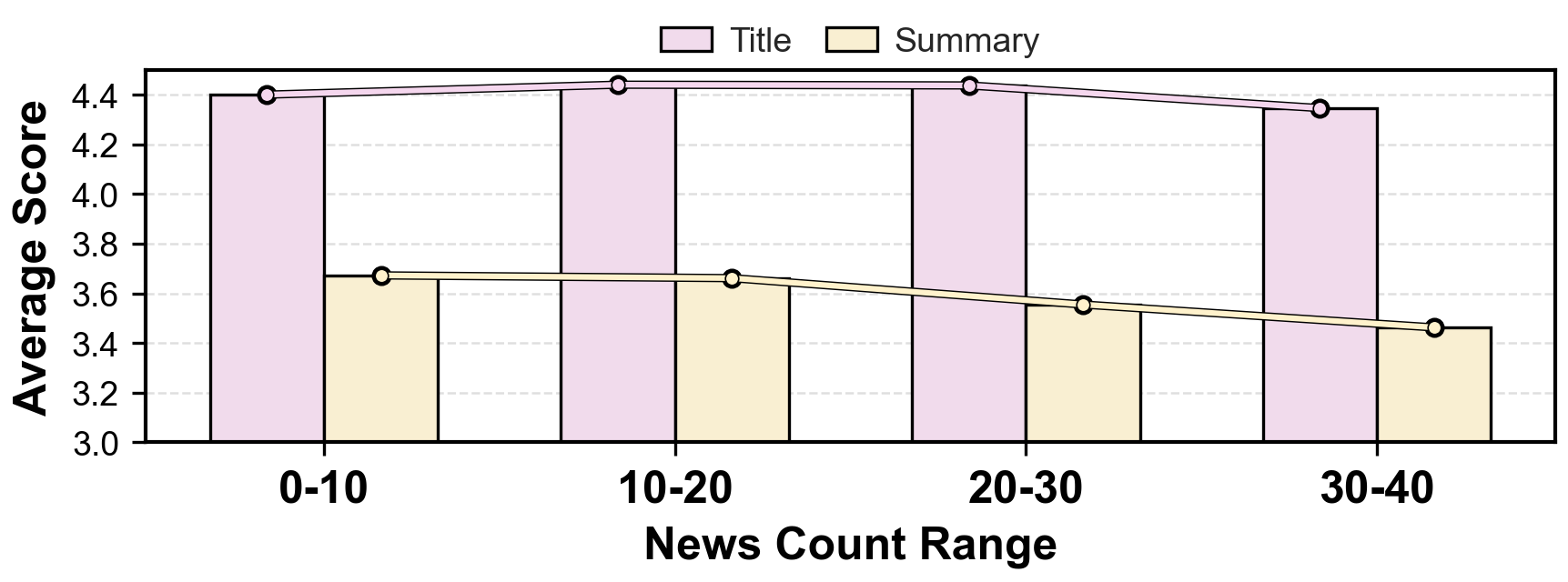}
    \caption{Average Title and Summary Scores by News Article Count Bins}
    \label{fig:rq-新闻数量}
\end{figure}

\begin{figure}[t]
    \centering
    \includegraphics[width=\linewidth]{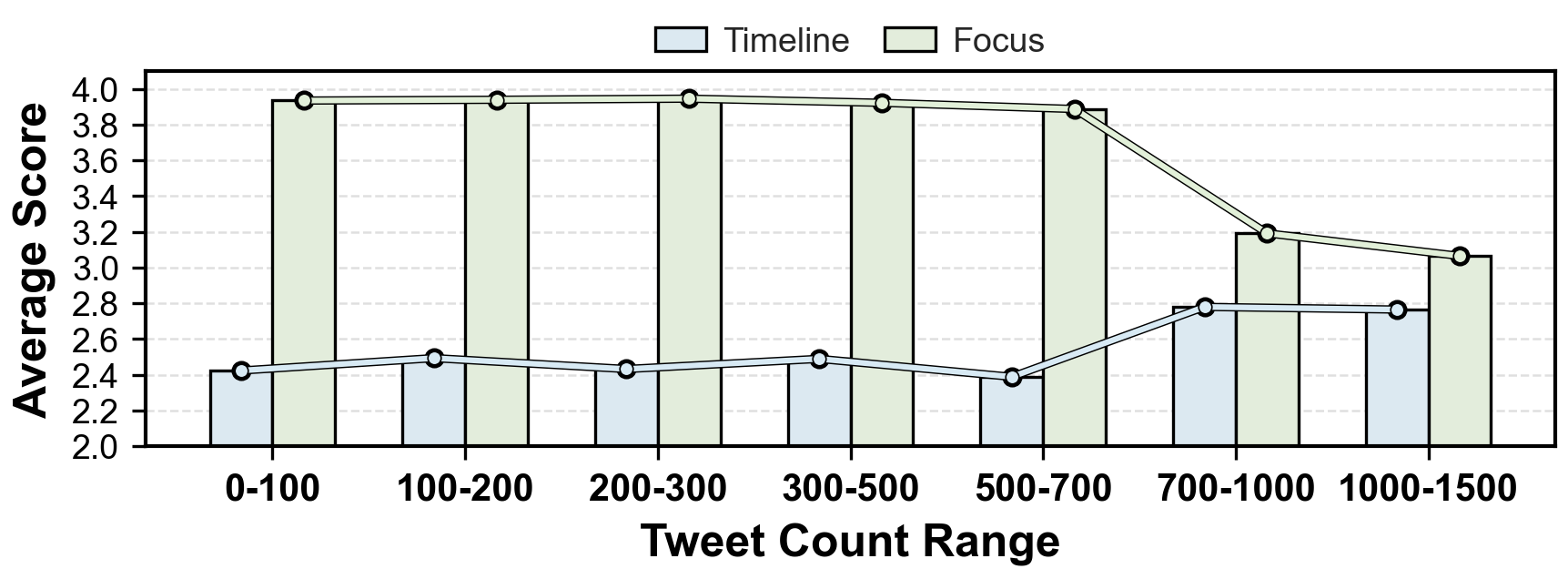}
    \caption{Average Timeline and Focus Scores by Tweet Count Bins}
    \label{fig:rq-推特数量}
\end{figure}

\begin{figure}[t]
    \centering
    \includegraphics[width=0.35\textwidth]{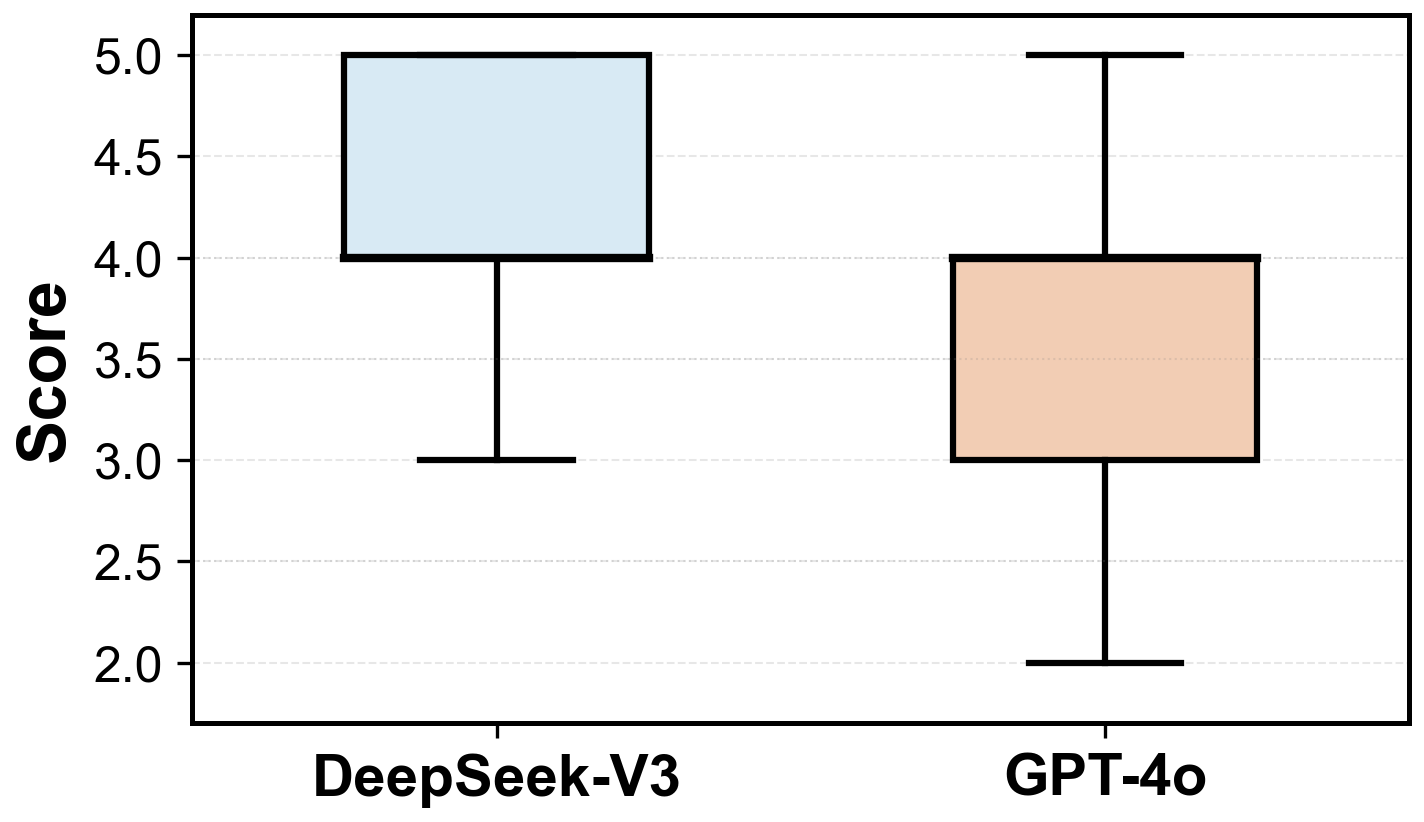}
    \caption{Systematic Scoring Bias Between DeepSeek-V3 and GPT-4o}
    \label{fig:rq-谁更严格}
\end{figure}

\begin{figure}[!t]
    \centering
    \includegraphics[width=0.45\textwidth]{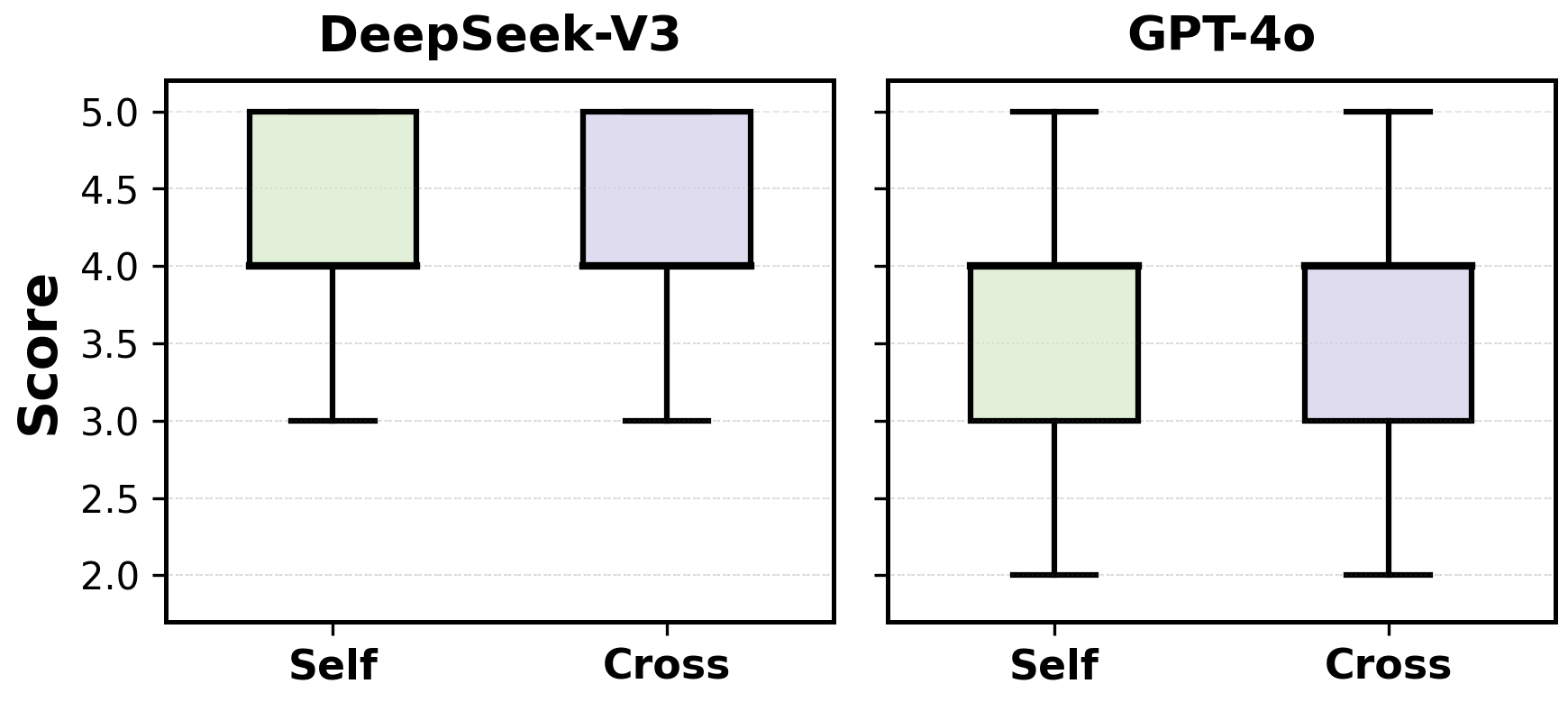}
    \caption{Comparative Analysis of Self-evaluation and Cross-evaluation Scores Between DeepSeek\-V3 and GPT\-4o.}
    \label{fig:rq-谁更自信}
\end{figure}

\subsubsection{LLMs Struggle with Information Overload and Multi-Document Synthesis.}
Our analysis reveals that report quality does not simply increase with source document volume, highlighting a critical limitation in the information overload and inefficient multi-document synthesis of current LLMs~\citep{wang2024rkld}.
Since Title and Summary generation primarily relies on news articles while Timeline and Focus depend on social media data, we analyze their correlations separately.

\textbf{News Articles:} 
Table~\ref{tab:news-tweet_count} shows a negative correlation between article count and the scores for both the Title and Summary, suggesting information overload. Figure~\ref{fig:rq-新闻数量} identifies the optimal range: 10-20 articles. 

\textbf{Social Media:} 
Table~\ref{tab:news-tweet_count} shows opposing effects: Timeline accuracy benefits from more tweets ($r=+0.21, p<0.001$), while Event Focus suffers dramatically ($r=-0.44, p<0.001$). 
Figure~\ref{fig:rq-推特数量} reveals a critical threshold approximately 700 tweets where this trade-off becomes pronounced.

\subsection{Evaluator Bias Analysis} 

\subsubsection{LLM Evaluators Display Inherent Scoring Biases.}
DeepSeek-V3 consistently assigns higher scores than GPT-4o for identical reports (average: 4.02 vs. 3.62, $p < 0.001$). 
Figure~\ref{fig:rq-谁更严格} visualizes this bias: GPT-4o shows lower median scores with wider distribution, indicating more stricter criteria and better discrimination, whereas DeepSeek-V3's scores cluster higher with less variance. 
This systematic difference proves that absolute scores from different LLM evaluators are not directly comparable, highlighting the need for score normalization or calibration in practical applications.

\subsubsection{LLM Evaluators Show Strong Objectivity with Negligible Self-Evaluation Bias.}
To investigate self-evaluation bias, we compare the scores an agent assigns to its own reports ("Self-Evaluation") versus those from other models ("Cross-Evaluation"). 
Our results (Figure~\ref{fig:rq-谁更自信}) show that while self-evaluation biases are statistically significant, their practical impact is negligible: DeepSeek-V3 shows a minor self-preference (+0.03 average score, $p < 0.001$), while GPT-4o exhibits slight self-criticism (-0.02 average score, $p < 0.05$). These differences represent less than 1\% of the rating scale.
This key finding confirms a high degree of objectivity in these LLM evaluators, enhancing the credibility and practical viability of LLM-based evaluation systems.


\section{Related Work}
\subsection{Multi-document Summarization}
Multi Document Summarization (MDS) generates comprehensive summaries from document collections on the same topic~\citep{Liu2024SumSurveyAA}, with applications in news extraction, social media mining, and review analysis~\citep{Bilal2022TemplatebasedAM,angelidis-lapata-2018-summarizing,Nallapati2016AbstractiveTS}. Research explores both extractive~\citep{Rossiello2017CentroidbasedTS,Mao2020MultidocumentSW,Zheng2019SubtopicdrivenMS} and abstractive approaches~\citep{chen2024essential,Ye2024GlobeSummAC,Liu2024SumSurveyAA,Cho2022TowardUT,Liu2022LeveragingLI}.

However, \taskname~differs fundamentally from MDS: (1) it generates structured reports with predefined sections rather than single paragraphs, and (2) it integrates both formal news and noisy social media posts. While MDS evaluation relies on n-gram metrics (ROUGE, BLEU)~\citep{bai-etal-2024-longbench,An2023LEvalIS,fabbri2019multi}, these are inadequate for long-form structured content~\citep{Celikyilmaz2020EvaluationOT,Krishna2021HurdlesTP}. Similar challenges exist in open-ended text generation tasks~\citep{ni2024xl,PicazoSanchez2024AnalysingTI}, where gold references are absent and human evaluation suffers from expertise limitations and subjectivity~\citep{Xu2023ACE}.

\subsection{Text Generation Evaluation}
LLM-based evaluation has recently emerged as a promising solution to the limitations of traditional metrics, demonstrating a strong correlation with human judgment while offering superior reproducibility, speed, and cost-effectiveness~\citep{Shen2023LargeLM,liu2025surveytransformercontextextension,qin2024large,chen2025towards,wang2023chatgpt,li2023loogle,chiang2023can,zhuang2023through}. 
A variety of strategies have been developed. For reference-free evaluation, methods employ techniques like chain-of-thought prompting or proxy question-answering~\citep{liu2023g,tan2024proxyqa}. Other research focuses on creating benchmarks to evaluate specific attributes, such as factual consistency~\citep{luo2023chatgpt}, instruction following~\citep{An2023LEvalIS,li2023loogle}, response alignment~\citep{zheng2024judging}, and even leveraging multi-agent systems for evaluation~\citep{wu2023large}.

Building on these advances, our \evaluationname~framework employs LLMs as intelligent agents, simulating expert evaluation by using generated reports as contextual background and applying 5-point Likert scale scoring tailored to \taskname~'s unique requirements.
\section{Conclusion}
In this paper, we address the critical inefficiency of manual online public opinion reporting.
To tackle this, we introduce three core contributions: the \taskname~task for automated report generation; \datasetname, the first multi-source benchmark to support it; and \evaluationname, a reliable agent-based evaluation framework achieving strong human correlation. 
Our experiments establish strong baselines and reveal key challenges, such as complex temporal reasoning and systematic biases inherent in different LLM evaluators. 
We believe this work not only provides practical guidance for public opinion management but also serves as a valuable resource for related NLP tasks like multi-document summarization and event extraction.

\bibliography{custom}

\appendix
\section{Preliminaries: The Structure of an Online Public Opinion Report}
\subsection{Event Timeline} \label{app:timeline}

Analyzing the public opinion lifecycle is strategically vital, shifting crisis management from reactive to proactive by enabling stage-specific guidance via the Event Timeline section~\citep{Ren2024SimulationOP,Yang2025ForwardingIS}. 
While established theories often divide the lifecycle into four phases (Incubation, Outbreak, Diffusion, Decline)~\citep{Zhang2021HowTR}, the sudden and fast-spreading nature of crises typically merges the Outbreak and Diffusion stages into a single, intense Peak Period, leading us to adopt a three-phase timeline: \textbf{Incubation}, \textbf{Peak}, and \textbf{Decline}.

\textbf{The Incubation Period} features discussion limited to directly affected stakeholders, making detection difficult without specialized monitoring~\citep{Zeng2021ACF,yang2016analysis}.
Despite low activity, these topics possess significant eruption potential—a controversial sub-event can rapidly trigger widespread attention~\citep{Wang2019StudiesOA,Lazebnik2023TemporalGA}. This period thus serves as a critical early-warning phase for anticipating crisis development~\citep{Ren2024SimulationOP,Dui2024SocialMP}.

\textbf{The Peak Period} exhibits exponential growth in attention, participation, volume, and velocity~\citep{Liu2021MultistageIP}. 
This expansion broadens scope and deepens complexity through derivative sub-events like official announcements and public controversies~\citep{Xing2023ExaminingTO}. 
The phase is a critical window for shaping public opinion, as it forges the public's long-term perception of the event~\citep{Yang2025ForwardingIS,Wang2020IdentifyingCO}.

\textbf{The Decline Period} shows diminishing public interest as focus shifts to newer events~\citep{Yang2025ForwardingIS,Liu2025TheMC}. 
Discussion reverts to directly affected stakeholders, where unresolved issues persist and can reactivate under specific conditions~\citep{He2018ResearchOP,Ren2024SimulationOP}. 
Thus, rather than a final resolution, this period marks a decline in widespread attention that leaves a lasting reputational impact~\citep{Mu2023IPSOLSTMHM,Jiang2021NetworkPO}.

\subsection{Event Focus} \label{app:focus}
During the Peak Period, an event triggers exponential growth in online discussions~\citep{Ren2024SimulationOP}, leading to topic polarization, rumor surges, and emotional instability~\citep{Hui2022ResearchOT,Mu2023IPSOLSTMHM}. 
Emotional contagion drives this growth, with anger fueling sharing and anxiety driving information-seeking~\citep{Khalil2024ExploringTP,Li2021AnAO}. 
This volatile mix necessitates multi-perspective analysis beyond event timelines~\citep{Mashayekhi2023MicroblogTD,Liu2023NetworkPO}, making group-specific analysis essential for prediction and intervention.

\subsection{Event Suggestions} \label{app:suggestions}
The Event Suggestions section bridges analysis and action by translating insights into actionable recommendations~\citep{Liu2023CrisisME,Zheng2017ExploringTR}, accelerating decision-making to mitigate public anxiety and mistrust arising from delayed responses~\citep{Wu2025HowTA,Utz2009OnlineRS}. 
Professional guidance enables prompt communication that fosters rational public response~\citep{Huang2024TheEC,Chen2013ExaminingTM}, whereas its absence renders analysis unactionable and hinders effective resource allocation and communication~\citep{Yang2025WarmthAC}.

\section{\datasetname~Collection Details}\label{app:collection_details}
\subsection{Crisis Event Identification}\label{app:event_sources}
The goal of our event collection is to create a large and diverse set of crisis events that have received sufficient media coverage and generated widespread social media discussion. 
We gather events from the two high-quality platforms:

\begin{itemize}
    \item \textbf{EM-DAT International Disaster Database:} Maintained by the Center for Research on the Epidemiology of Disasters (CRED), providing standardized records with verified impact metrics.

    \item \textbf{Wikipedia's Curated Disaster Lists:} These lists are community-vetted and often link to well-referenced articles, making them a reliable source. Specific lists we utilize include, but are not limited to earthquakes\footnote{\url{https://en.wikipedia.org/wiki/Lists_of_earthquakes}}, floods\footnote{\url{https://en.wikipedia.org/wiki/List_of_floods_in_Europe}}, and wildfires\footnote{\url{https://en.wikipedia.org/wiki/List_of_wildfires}}.
\end{itemize}

\subsection{Document Collection Procedures}\label{app:document_collection}
For each of the 463 events in our corpus, we execute two parallel streams to gather high-quality news articles and comprehensive social media posts.

\paragraph{News Article Collection}
To ensure the collected news articles are both high-quality and relevant, we implement a two-phase strategy:
\begin{enumerate}
    \item \textbf{Initial Crawling from Vetted Sources:} In the first phase, we crawl news articles directly from the "References" section of each event's official Wikipedia page (e.g., the page for the "2025 Table Mountain fire"\footnote{\url{https://en.wikipedia.org/wiki/2025_Table_Mountain_fire}}). We leverage these community-vetted citations as a reliable first-pass quality filter.
    \item \textbf{Relevance Reranking:} In the refinement phase, we employ the BM25 retriever~\citep{Lin2021PyseriniAP} to rerank the collected articles. The reranking is based on each article's relevance to event-specific keywords (e.g., "2025\_Table\_Mountain\_fire", "2025\_Israel\_West\_Bank\_fires"), ensuring that the final set of articles is tightly focused on the event.
\end{enumerate}

\subsection{LLM Selection for Annotation Framework}\label{app:llm_selection}
\paragraph{Experimental Setup}
We evaluate a set of frontier LLMs on the Social Media Annotation task. We use the pre-labeled data from the CrisisLexT26 dataset~\citep{olteanu2015expect} as the ground truth for this evaluation. Each model is prompted to classify the author type of tweets from the dataset.

\paragraph{Results and Analysis}
\begin{table}[ht]
\centering
\small 
\caption{Performance comparison of different models on the Social Media Annotation. The best result is highlighted in bold.}
\label{tab:llm_comparison}
\begin{tabular}{lc}
\toprule
\textbf{Model} & \textbf{Macro-F1} \\
\midrule
Gemini 2.5 Pro & \textbf{0.8053} \\
DeepSeek-V3 & 0.8012 \\
GPT-4o & 0.8009 \\
DeepSeek-R1 & 0.7934 \\
GPT-4o-mini & 0.7800 \\
Grok-3 Reasoner & 0.6235 \\
Claude 3.7 & 0.6127 \\
\bottomrule
\end{tabular}
\end{table}
As shown in Table~\ref{tab:llm_comparison}, gpt-4o-mini demonstrates the optimal balance between annotation quality and cost, so we select it as the primary model for our protocol-guided annotation tasks.


\subsubsection{Reference and Social Media Annotation}\label{app:reference_and-social-media_prompt}
\begin{tcolorbox}[
    breakable,
    title=Prompt Template for Factual Attribute Extraction,
    colback=white,
    colframe=black!75,
    boxrule=0.5pt,
    left=4pt, right=4pt, top=2pt, bottom=2pt,
    arc=2mm,
    fonttitle=\bfseries,
]

\begin{center}\textbf{[SYSTEM PROMPT]}\end{center}

\paragraph{1. Role and Goal}
You are a meticulous data analyst. Your task is to read the provided source text about a single public opinion event and extract specific factual attributes. Your output \textbf{MUST} be a single, valid JSON object and nothing else.

\paragraph{2. Field-by-Field Annotation Guideline}
You must populate the following fields based on the source text. If a piece of information cannot be found in the text, its value \textbf{MUST} be \texttt{null}.

\begin{itemize}
    \item \textbf{time}:
    \begin{itemize}
        \item \texttt{"development"}: A concise sentence describing the origin and formation of the event.
        \item \texttt{"duration\_days"}: The total duration of the event in days (integer).
        \item \texttt{"start\_day"}: The specific start date in "YYYY-MM-DD" format.
        \item[...]
    \end{itemize}
    \item \textbf{location}:
    \begin{itemize}
        \item \texttt{"country"}: The primary country where the crisis occurred.
        \item \texttt{"spread"}: A string listing all significant regions or countries affected.
        \item[...]
    \end{itemize}
    \item \textbf{impact}:
    \begin{itemize}
        \item \texttt{"fatalities"}: The total number of deaths (integer).
        \item \texttt{"damage"}: A string describing the estimated economic damage.
        \item[...]
    \end{itemize}
    \item[] ... (and so on for \texttt{categorization}, \texttt{involved\_parties}, and \texttt{causes})
\end{itemize}

\paragraph{3. Few-Shot Example}
\textit{This section provides a concrete input-output pair to guide the model.}

\textbf{Example Input Text}:
\begin{lstlisting}[style=myjson]
...source text about Tropical Storm Amanda...|

\textbf{Example Output JSON (Snippet)}:
\begin{lstlisting}[style=myjson]
{
  "time": {
    "development": "Tropical Storm Amanda developed...",
    "duration_days": 14,
    "start_day": "2020-05-30"
  },
  "impact": {
    "fatalities": 46,
    "damage": "$865 million (2020 USD)",
    ...
  },
  ...
}
\end{lstlisting}

\hrulefill 

\begin{center}\textbf{[INPUT TEXT]}\end{center}
\verb|{source_text_to_be_annotated}|

\end{tcolorbox}


\begin{tcolorbox}[
    breakable,
    title=Prompt Template for Social Media Source Classification,
    colback=white,
    colframe=black!75,
    boxrule=0.5pt,
    left=4pt, right=4pt, top=2pt, bottom=2pt,
    arc=2mm,
    fonttitle=\bfseries,
]

\begin{center}\textbf{[SYSTEM PROMPT]}\end{center}

\paragraph{1. Role and Goal}
You are an expert public opinion classifier. Your task is to analyze the provided tweet and classify its author's identity into one of two categories: \texttt{Netizens} or \texttt{Authoritative Institutions}.

\paragraph{2. Category Definitions and Rules}
You must classify the source based on the following definitions. Your final response \textbf{MUST} be only one of the two category names and nothing else.
\begin{itemize}
    \item \textbf{Netizens:} The language is typically informal. The content often expresses personal opinions, emotions, or provides eyewitness accounts.
    \item \textbf{Authoritative Institutions:} The language is typically formal or neutral. This category includes media outlets, government agencies, NGOs, etc.
\end{itemize}

\paragraph{3. Few-Shot Examples}
\textit{This section provides examples to guide the model.}

\textbf{Example 1 (Netizen):}
\begin{itemize}
    \item \textit{Input Tweet:} \verb|{example_netizen_tweet}|
    \item \textit{Output:} \texttt{Netizens}
\end{itemize}

\textbf{Example 2 (Authoritative Institution):}
\begin{itemize}
    \item \textit{Input Tweet:} \verb|{example_institution_tweet}|
    \item \textit{Output:} \texttt{Authoritative Institutions}
\end{itemize}

\hrulefill 

\begin{center}\textbf{[INPUT TWEET]}\end{center}
\verb|{tweet_text}|

\end{tcolorbox}

\subsection{Timeline Annotation}\label{app:timeline_tool}
\subsubsection{Timeline Annotation Protocol} \label{app:timeline_protocol}
To ensure the highest reliability for the timeline annotations, we employ a rigorous, multi-stage process. This protocol, conducted by six in-house experts (Master’s and PhD students familiar with public opinion lifecycles), proceeds as follows:

\begin{itemize}
    \item The 463 events are evenly divided among three groups, with two annotators per group.
    \item Both annotators within each group work independently on the same set of assigned events.
    \item The two partners then compare and consolidate their results, and are required to discuss every disagreement until a consensus is reached.
    \item For the rare cases where a consensus cannot be reached, a senior researcher performs a final adjudication. If ambiguity persists, the event is discarded to ensure data integrity.
\end{itemize}
\subsubsection{Annotation Interfaces}

\begin{figure*}[h!]
    \centering
    \includegraphics[width=0.75\textwidth]{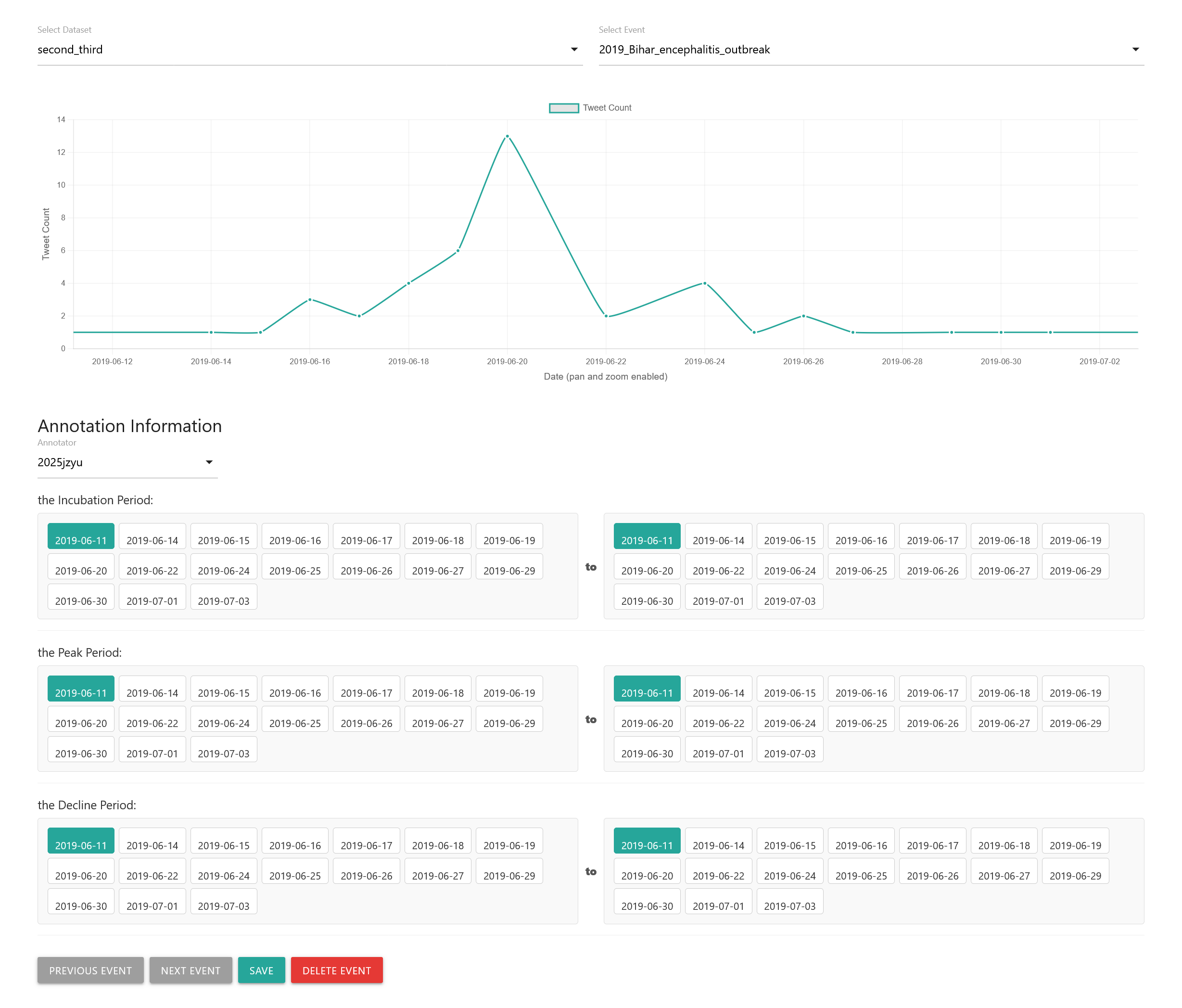}
    \caption{Screenshot of the Timeline annotation interface for the Incubation Period, the Peak Period, and the Decline Period.}
    \label{fig:timeline_annotation}
\end{figure*}

\section{Detailed Scoring Guidelines}\label{app:evaluation_criteria}
The following sections provide the detailed scoring criteria for all 15 evaluation dimensions. Adopting a methodology similar to that of ~\citet{kocmi2023large}, each dimension is rated on a 5-point Likert scale, where a score of 1 indicates an unacceptable generation and 5 represents an excellent one.

\subsection{Fact-Checker Tool: Scoring Guidelines} \label{app:fact_checker_criteria}

\subsubsection{Scoring Guideline for Event Title}
\label{app:title_guideline}

\begin{tcolorbox}[
    breakable,  
    title=Scoring Guideline for Event Title, 
    colback=white,  
    colframe=black!75, 
    boxrule=0.5pt, 
    left=4pt, 
    right=4pt, 
    top=2pt, 
    bottom=2pt, 
    arc=2mm, 
    fonttitle=\bfseries, 
]

The quality of the "Event\_Title" is rated on a scale of 1 to 5. Your evaluation should assess to what extent the title incorporates the official "event name" and relevant "keywords" to be clear, specific, and instantly recognizable.

\begin{description}
    \item[\textbf{Score 5 (Excellent):}] The title perfectly incorporates the official event name (or its recognized alternative) and key keywords in a clear and coherent manner, precisely and unambiguously identifying the crisis event.
    
    \item[\textbf{Score 4 (Good):}] The title clearly references the event name and relevant keywords, allowing readers to readily identify the crisis, though there may be minor room for improvement.
    
    \item[\textbf{Score 3 (Fair):}] The title partially mentions the event name or a few keywords, broadly pointing to the correct crisis but lacking clarity and completeness.
    
    \item[\textbf{Score 2 (Poor):}] The title provides only minimal or vague hints related to the event, leaving the specific crisis unclear to the reader.
    
    \item[\textbf{Score 1 (Unacceptable):}] The title completely fails to mention the event name or any relevant keywords, providing no clear indication of the crisis.
\end{description}

\end{tcolorbox}

\subsubsection{Scoring Guideline for Event Summary}
\label{app:summary_guideline}

\begin{tcolorbox}[
    breakable,
    title=Scoring Guidelines for Event Summary,
    colback=white,
    colframe=black!75,
    boxrule=0.5pt,
    left=4pt,
    right=4pt,
    top=2pt,
    bottom=2pt,
    arc=2mm,
    fonttitle=\bfseries,
]

This section details the five scoring dimensions for the \textit{Event Summary}. Each dimension is rated on a 5-point scale based on its correspondence with the source materials.

\begin{enumerate}
    \item \textbf{Event Nature (Categorization):} This criterion judges if the summary accurately describes the event's type (e.g., social issue, natural disaster, corporate crisis).
    
    \begin{description}
        \item[\textbf{Score 5 (Excellent):}] Accurately and comprehensively categorizes the event, clearly describing its core activities and nature in line with the source.
        \item[\textbf{Score 4 (Good):}] Clearly mentions the event type and covers most of its essential aspects as described in the source.
        \item[\textbf{Score 3 (Fair):}] Mentions the event category but lacks depth or specific details that were present in the source.
        \item[\textbf{Score 2 (Poor):}] Vaguely hints at the event type without clarity or sufficient detail.
        \item[\textbf{Score 1 (Unacceptable):}] Completely fails to mention or misrepresents the event's type or essence.
    \end{description}

    \item \textbf{Time and Location:} This criterion assesses the accuracy and completeness of the temporal and geographical information provided.
    
    \begin{description}
        \item[\textbf{Score 5 (Excellent):}] Clearly and accurately specifies the time (e.g., start date) and location as detailed in the source, with full context if provided.
        \item[\textbf{Score 4 (Good):}] Provides a relatively accurate description of both time and location, consistent with the source.
        \item[\textbf{Score 3 (Fair):}] Partially mentions time and/or location from the source, but the information is incomplete or lacks precision.
        \item[\textbf{Score 2 (Poor):}] Mentions vague or unclear time/place (e.g., "recently", "in a certain country"), significantly different from the source.
        \item[\textbf{Score 1 (Unacceptable):}] Does not mention time or location, or provides completely incorrect information.
    \end{description}

    \item \textbf{Involved Parties:} This criterion evaluates if the summary correctly identifies the key actors and their roles.
    
    \begin{description}
        \item[\textbf{Score 5 (Excellent):}] Thoroughly and accurately lists all significant parties mentioned in the source, detailing their roles and, if specified in the source, their motivations/actions.
        \item[\textbf{Score 4 (Good):}] Identifies most main parties from the source and their roles, with some indication of their motivations or actions as per the source.
        \item[\textbf{Score 3 (Fair):}] Identifies some key parties and their basic roles, but misses important actors or details on their involvement/motivations.
        \item[\textbf{Score 2 (Poor):}] Vaguely mentions some parties without clarifying their roles or if they match the source.
        \item[\textbf{Score 1 (Unacceptable):}] No mention of involved parties, or completely misidentifies them.
    \end{description}
    
    \item \textbf{Causes:} This criterion judges if the summary accurately explains the triggers and underlying factors of the event.
    
    \begin{description}
        \item[\textbf{Score 5 (Excellent):}] Thoroughly and accurately analyzes the causes as detailed in the source, demonstrating a comprehensive understanding of the underlying factors described.
        \item[\textbf{Score 4 (Good):}] Clearly mentions the main causes as described in the source, with some analysis of background factors if present in the source.
        \item[\textbf{Score 3 (Fair):}] Provides a partial analysis of causes mentioned in the source, but lacks depth or misses key factors.
        \item[\textbf{Score 2 (Poor):}] Briefly mentions causes with very little detail or accuracy compared to the source.
        \item[\textbf{Score 1 (Unacceptable):}] No mention of causes, or attributes causes completely unrelated to the source.
    \end{description}
    
    \item \textbf{Impact:} This criterion assesses if the summary covers the short- and long-term consequences of the event.
    
    \begin{description}
        \item[\textbf{Score 5 (Excellent):}] Thoroughly and accurately describes the multi-faceted impacts (e.g., on individuals, communities, environment; short/long-term) as detailed in the source.
        \item[\textbf{Score 4 (Good):}] Mentions significant impacts (e.g., short-term and long-term if applicable) as per the source, with reasonable detail.
        \item[\textbf{Score 3 (Fair):}] Describes some impacts mentioned in the source but lacks depth, or fails to distinguish between types of impact (e.g., short/long-term) if the source did.
        \item[\textbf{Score 2 (Poor):}] Mentions vague or minimal impact that is not well-aligned with the source.
        \item[\textbf{Score 1 (Unacceptable):}] No mention of impact, or describes impacts completely unrelated to the source.
    \end{description}

\end{enumerate}

\end{tcolorbox}

\subsubsection{Scoring Guideline for Timeline: Date Accuracy}
\label{app:timeline_date_guideline}

\begin{tcolorbox}[
    breakable,
    title=Scoring Guideline for Date Accuracy,
    colback=white,
    colframe=black!75,
    boxrule=0.5pt,
    left=4pt, right=4pt, top=2pt, bottom=2pt,
    arc=2mm,
    fonttitle=\bfseries,
]

This criterion assesses the factual accuracy of the key dates in the \textit{Event\_Timeline}. The evaluation is based on four critical reference dates provided for the event's lifecycle: the start of the \textbf{Incubation Period}, the start of the \textbf{Peak Period}, and the start and end of the \textbf{Decline Period}.

The final 1-5 score is determined based on how many of these four critical dates are correctly reflected in the generated timeline. The mapping is as follows:

\begin{description}
    \item[\textbf{Score 5 (Excellent):}] The timeline correctly reflects or aligns with all \textbf{four} of the reference dates.
    
    \item[\textbf{Score 4 (Good):}] The timeline correctly reflects or aligns with \textbf{three} of the four reference dates.
    
    \item[\textbf{Score 3 (Fair):}] The timeline correctly reflects or aligns with \textbf{two} of the four reference dates.
    
    \item[\textbf{Score 2 (Poor):}] The timeline correctly reflects or aligns with only \textbf{one} of the four reference dates.
    
    \item[\textbf{Score 1 (Unacceptable):}] The timeline fails to correctly reflect \textbf{any} of the four reference dates.
\end{description}

\end{tcolorbox}

\subsection{Opinion-Miner Tool: Scoring Guidelines} \label{app:opinion_miner_criteria}
\subsubsection{Scoring Guideline for Timeline: Sub-Events Coverage}
\label{app:timeline_subevent_guideline}

\begin{tcolorbox}[
    breakable,
    title=Scoring Guideline for Sub-Events Coverage,
    colback=white,
    colframe=black!75,
    boxrule=0.5pt,
    left=4pt, right=4pt, top=2pt, bottom=2pt,
    arc=2mm,
    fonttitle=\bfseries,
]

This criterion assesses how well the \textit{Event\_Timeline} captures the key sub-events that occurred during each phase (Incubation, Peak, Decline). The evaluation is based on a comparison against the source social media posts~($Y_i$).

The evaluation follows a two-step process:
\begin{enumerate}
    \item \textbf{Ground Truth Identification:} Key sub-events for each phase are identified from the provided social media context, focusing on recurring topics, emotional peaks, public reactions, and official statements.
    \item \textbf{Comparison:} The generated \textit{Event\_Timeline} text is then compared against these identified sub-events to assess its coverage and accuracy for each stage.
\end{enumerate}
The final score is assigned based on the following standards:

\begin{description}
    \item[\textbf{Score 5 (Excellent):}] Fully captures all major sub-events for all periods, aligning well with the key happenings suggested by the reference social media context.
    
    \item[\textbf{Score 4 (Good):}] Has good coverage of sub-events across the periods, with only minor omissions or lack of detail.
    
    \item[\textbf{Score 3 (Fair):}] Covers sub-events for at least two periods with reasonable accuracy, but details are lacking in depth for one or more periods.
    
    \item[\textbf{Score 2 (Poor):}] Only partially covers one period’s sub-events and has significant omissions in others.
    
    \item[\textbf{Score 1 (Unacceptable):}] Fails to capture the key sub-events across all three periods based on the reference social media context.
\end{description}

\end{tcolorbox}

\subsubsection{Scoring Guideline for Event Focus}
\label{app:focus_guideline}

\begin{tcolorbox}[
    breakable,
    title=Scoring Guidelines for Event Focus,
    colback=white,
    colframe=black!75,
    boxrule=0.5pt,
    left=4pt, right=4pt, top=2pt, bottom=2pt,
    arc=2mm,
    fonttitle=\bfseries,
]

This section details the three scoring dimensions for the \textit{Event Focus}. The evaluation requires comparing the generated text against the viewpoints, statements, and sentiments identified from the source social media posts~($Y_i$). Each dimension is rated on a 5-point scale.

\begin{enumerate}
    \item \textbf{Core Topic \& Viewpoint Extraction:} This criterion assesses if the \textit{Event Focus} accurately captures the key topics and controversial debates from both Netizen and Authoritative Institution groups.

    \begin{description}
        \item[\textbf{Score 5 (Excellent):}] Fully and accurately captures all key topics and controversies for both groups, clearly distinguishing their respective viewpoints.
        \item[\textbf{Score 4 (Good):}] Covers most major topics and controversies accurately, with only minor omissions or misattributions between groups.
        \item[\textbf{Score 3 (Fair):}] Covers at least two key topics for each group, but some important debates or viewpoints are omitted or lack depth.
        \item[\textbf{Score 2 (Poor):}] Covers basic topics for only one of the two groups, while largely omitting key debates and the other group's perspective.
        \item[\textbf{Score 1 (Unacceptable):}] Fails to capture the core topics and debates, showing major discrepancies with the source content.
    \end{description}

    \item \textbf{Representative Statements Comparison:} This criterion evaluates if the \textit{Event Focus} includes specific statements that are truly representative of the viewpoints held by each group.

    \begin{description}
        \item[\textbf{Score 5 (Excellent):}] Includes comprehensive and highly representative statements for both groups, providing strong, illustrative evidence for their stances.
        \item[\textbf{Score 4 (Good):}] Includes mostly representative statements for both groups, with some good examples illustrating their key viewpoints.
        \item[\textbf{Score 3 (Fair):}] Includes a few representative statements, but they may lack balance between the groups or be only partially illustrative.
        \item[\textbf{Score 2 (Poor):}] Includes only vague or generic perspectives, lacking specific, representative statements.
        \item[\textbf{Score 1 (Unacceptable):}] Fails to include any specific, representative statements from either group.
    \end{description}

    \item \textbf{Emotional \& Motivational Analysis:} This criterion judges if the \textit{Event Focus} accurately identifies the emotional tone and underlying motivations for each group's viewpoints.
    
    \begin{description}
        \item[\textbf{Score 5 (Excellent):}] Precisely captures the emotional tone and provides insightful analysis of the underlying motivations for each viewpoint, fully aligning with the source.
        \item[\textbf{Score 4 (Good):}] Provides a solid analysis of emotions and motivations, with only minor omissions or lack of nuance.
        \item[\textbf{Score 3 (Fair):}] Captures the general sentiment (e.g., positive/negative) correctly but lacks depth in analyzing more complex emotions or motivations.
        \item[\textbf{Score 2 (Poor):}] Provides a vague or partially inaccurate sentiment analysis, with little to no analysis of motivations.
        \item[\textbf{Score 1 (Unacceptable):}] Completely misrepresents the emotional tone and provides no valid analysis of motivations.
    \end{description}
\end{enumerate}

\end{tcolorbox}

\subsection{Solution Counselor Tool: Scoring Guidelines} \label{app:solution_counselor_criteria}
\subsubsection{Scoring Guideline for Event Suggestions}
\label{app:suggestions_guideline}

\begin{tcolorbox}[
    breakable,
    title=Scoring Guidelines for Event Suggestions,
    colback=white,
    colframe=black!75,
    boxrule=0.5pt,
    left=4pt, right=4pt, top=2pt, bottom=2pt,
    arc=2mm,
    fonttitle=\bfseries,
]

This section details the four scoring dimensions for the \textit{Event Suggestions}. Each dimension is rated on a 5-point scale.

\begin{enumerate}
    \item \textbf{Relevance \& Coverage:} This criterion evaluates whether the suggestions align with the event's core issues and cover all relevant stakeholders.
    
    \begin{description}
        \item[\textbf{Score 5 (Very High):}] Thoroughly addresses both primary and secondary issues; caters well to the needs of all stakeholders involved.
        \item[\textbf{Score 4 (High):}] Closely addresses the core controversies and accounts for most major stakeholders.
        \item[\textbf{Score 3 (Medium):}] Covers most primary concerns but offers limited attention to secondary issues or certain groups.
        \item[\textbf{Score 2 (Low):}] Addresses some issues but misses critical focal points; neglects important stakeholders.
        \item[\textbf{Score 1 (Very Low):}] Largely unrelated to the main controversies or focuses on only a small subset of stakeholders/topics.
    \end{description}

    \item \textbf{Effectiveness \& Feasibility:} This criterion assesses the practicality of the suggestions and their potential to resolve the key issues.
    
    \begin{description}
        \item[\textbf{Score 5 (Very High):}] Includes comprehensive planning for both short- and long-term resolution; highly likely to significantly resolve or reduce the problem.
        \item[\textbf{Score 4 (High):}] Presents a clear solution path with well-thought-out steps; effectively mitigates core controversies.
        \item[\textbf{Score 3 (Medium):}] Partially solves key issues; shows preliminary feasibility but needs further elaboration.
        \item[\textbf{Score 2 (Low):}] Contains a basic idea but lacks operational details; implementation and impact are questionable.
        \item[\textbf{Score 1 (Very Low):}] Difficult to implement; offers little or no real impact on the core problems.
    \end{description}

    \item \textbf{Emotional Guidance \& Trust-Building:} This criterion examines how the suggestions propose to manage public sentiment and repair trust.
    
    \begin{description}
        \item[\textbf{Score 5 (Very High):}] Implements a multi-layered approach to emotional support and trust repair, with ongoing measures to sustain improvements.
        \item[\textbf{Score 4 (High):}] Offers a systematic emotional management framework; fosters positive dialogue and helps rebuild trust among key parties.
        \item[\textbf{Score 3 (Medium):}] Addresses public emotions with basic strategies; provides an initial approach to easing tensions.
        \item[\textbf{Score 2 (Low):}] Briefly mentions emotion or trust but does not propose concrete steps or engagement plans.
        \item[\textbf{Score 1 (Very Low):}] Ignores emotional factors and trust repair; lacks meaningful communication strategy.
    \end{description}
    
    \item \textbf{Innovation:} This criterion assesses whether the suggestions offer novel or forward-thinking approaches beyond standard practices.
    
    \begin{description}
        \item[\textbf{Score 5 (Very High):}] Demonstrates a disruptive or revolutionary perspective, providing an entirely fresh angle to tackle the issues.
        \item[\textbf{Score 4 (High):}] Proposes clearly innovative, forward-thinking solutions; shows strong originality.
        \item[\textbf{Score 3 (Medium):}] Displays some creative elements or novel concepts but lacks a fully developed, breakthrough approach.
        \item[\textbf{Score 2 (Low):}] Makes slight improvements within a conventional framework, offering limited creativity.
        \item[\textbf{Score 1 (Very Low):}] Strictly follows outdated or formulaic methods, showing no new ideas.
    \end{description}
\end{enumerate}

\end{tcolorbox}

\section{Detailed Prompt Templates}\label{app:prompt_templates}
\subsection{Prompts for Generation Strategies} \label{app:prompts-generation}
This section provides the detailed prompt templates used for the two generation strategies (end-to-end and modular) described in our experiments. 

\subsubsection{Prompt for End-to-end Strategy} \label{app:prompt-end-to-end}
\begin{tcolorbox}[
    breakable,
    title=Prompt Template for End-to-end Generation Approach,
    colback=white,
    colframe=black!75,
    boxrule=0.5pt,
    left=4pt, right=4pt, top=2pt, bottom=2pt,
    arc=2mm,
    fonttitle=\bfseries,
]

\begin{center}\textbf{[SYSTEM PROMPT]}\end{center}

\paragraph{1. Role and Goal}
You are an expert public opinion analyst. Your primary task is to analyze the provided context (news articles) and input (tweets) to generate a complete, structured public opinion report in a single pass. The output must be a valid JSON object.

\paragraph{2. Field-by-Field Generation Instructions}
You must generate content for all five report sections, adhering to the following key guidelines:

\begin{itemize}
    \item \textbf{Event\_Title:} Generate a concise and accurate event title.
    
    \item \textbf{Event\_Summary:} Generate a detailed summary covering the five core dimensions (Crisis Type, Time/Location, Cause, Impact, etc.).
    
    \item \textbf{Event\_Focus:} Classify tweets (Netizens/Authoritative Institutions), perform topic clustering and sentiment analysis, and extract 2-3 key viewpoints for each group.
    
    \item[] ... (and so on for \textit{Event\_Timeline} and \textit{Event\_Suggestions}, with their respective constraints).
\end{itemize}

\paragraph{3. Few-Shot Examples}
\textit{This section illustrates the expected style and depth for each field.}

\begin{lstlisting}[style=myjson]
--- Content Examples for "Event_Summary" field ---
{summary_style_examples}

--- Content Examples for "Event_Focus" field ---
{focus_style_examples}

... (and so on for other fields)
\end{lstlisting}

\hrulefill 

\begin{center}\textbf{[TASK DATA]}\end{center}

\textbf{--- News Data (Context) Below ---}
\texttt{<input>}

\vspace{1em}

\textbf{--- Twitter Data (Input) Below ---}
\texttt{<input>}

\end{tcolorbox}

\subsubsection{Prompts for Modular Strategy} \label{app:prompts-modular}

\begin{tcolorbox}[
    breakable,
    title=Prompt Template for Event Title Generation,
    colback=white,
    colframe=black!75,
    boxrule=0.5pt,
    left=4pt, right=4pt, top=2pt, bottom=2pt,
    arc=2mm,
    fonttitle=\bfseries,
]
\paragraph{1. Role and Goal}
You are an expert public opinion analyst. Your task is to generate a concise, neutral, and highly descriptive title based on the news content provided below.

\paragraph{2. Field-by-Field Generation Instructions} 
\begin{itemize}
    \item \textbf{Event\_Title:} The title should capture the core essence of the event in a single, clear phrase.
\end{itemize}

\paragraph{3. Few-Shot Examples}
\begin{itemize}
    \item \textbf{Input:} \verb|<input_example_1>|
    \item \textbf{Output Title:} \verb|<title_example_1>|
\end{itemize}

\hrulefill 

\textbf{--- News Content (Context) Below ---}

\texttt{<input>}

\end{tcolorbox}

\begin{tcolorbox}[
    breakable,
    title=Prompt Template for Event Summary Generation,
    colback=white,
    colframe=black!75,
    boxrule=0.5pt,
    left=4pt, right=4pt, top=2pt, bottom=2pt,
    arc=2mm,
    fonttitle=\bfseries,
]
\paragraph{1. Role and Goal}
You are an expert public opinion analyst. Your task is to generate a concise and comprehensive event summary (around 100-150 words) based on the news content provided below.

\paragraph{2. Field-by-Field Generation Instructions}
The "Event\_Summary" should be a single block of text covering the following key crisis components:
\begin{itemize}
    \item \textbf{Type of Crisis}: Identify whether it is a natural, social, or man-made crisis.
    \item \textbf{Event and Location}: Describe the event and specify where it took place.
    \item \textbf{Cause of the Crisis}: Identify the underlying causes that led to the crisis.
    \item \textbf{Impact}: Describe the impact the crisis has had.
    \item \textbf{Time}: Clearly state when the event occurred.
\end{itemize}

\paragraph{3. Few-Shot Examples}
\begin{itemize}
    \item \textbf{Input:} \verb|<input_example_1>|
    \item \textbf{Output Summary:} \verb|<summary_example_1>|
\end{itemize}

\hrulefill

\textbf{--- News Content (Context) Below ---}

\texttt{<input>}

\end{tcolorbox}

\begin{tcolorbox}[
    breakable,
    title=Prompt Template for Event Timeline Generation,
    colback=white,
    colframe=black!75,
    boxrule=0.5pt,
    left=4pt, right=4pt, top=2pt, bottom=2pt,
    arc=2mm,
    fonttitle=\bfseries,
]
\paragraph{1. Role and Goal}
You are an expert public opinion analyst. Your task is to generate a comprehensive analysis of the event's timeline based on the provided news articles and social media data. The timeline \textbf{MUST} be structured into three distinct phases: the \textbf{Incubation Period}, the \textbf{Peak Period}, and the \textbf{Decline Period}.

\paragraph{2. Field-by-Field Generation Instructions}
For each of the three phases, you must provide a detailed narrative paragraph that includes the following key aspects, based on the provided data (e.g., volume of posts, news content, comments):
\begin{itemize}
    \item \textbf{Key Sub-Events:} Describe the specific milestones or key developments that occurred during that period.
    \item \textbf{Evolution Analysis:} Explain how public opinion shifted or evolved, capturing the dynamics of the event during that phase.
\end{itemize}

\paragraph{3. Few-Shot Examples}
\begin{itemize}
    \item \textbf{Input:} \verb|<input_example_1>|
    \item \textbf{Output Timeline:} \verb|<timeline_example_1>|
\end{itemize}

\hrulefill

\textbf{--- Twitter Data Below ---}

\texttt{<input>}

\end{tcolorbox}

\begin{tcolorbox}[
    breakable,
    title=Prompt Template for Event Focus Generation,
    colback=white,
    colframe=black!75,
    boxrule=0.5pt,
    left=4pt, right=4pt, top=2pt, bottom=2pt,
    arc=2mm,
    fonttitle=\bfseries,
]
\paragraph{1. Role and Goal}
You are an expert public opinion analyst. Your task is to generate a detailed analysis of the \textit{Event Focus} by comparing the viewpoints of two distinct groups—\textbf{Netizens} and \textbf{Authoritative Institutions}—based on the provided social media data.

\paragraph{2. Generation Instructions}
Your analysis must be structured by first presenting the findings for \textbf{Netizens}, followed by the findings for \textbf{Authoritative Institutions}. For each group, you must provide the following three components:
\begin{enumerate}
    \item \textbf{Core Topics:} A summary of the major topics discussed by this group.
    \item \textbf{Overall Emotional Tone:} The dominant sentiment (e.g., positive, negative, neutral) of the group's conversation.
    \item \textbf{Key Viewpoints:} 2-3 key viewpoints that capture the essence of the group's stance, using direct excerpts from the input text where possible.
\end{enumerate}

\paragraph{3. Few-Shot Examples}
\begin{itemize}
    \item \textbf{Input:} \verb|<input_example_1>|
    \item \textbf{Output Focus Analysis:} \verb|<focus_analysis_example_1>|
\end{itemize}

\hrulefill


\textbf{--- Twitter Data Below ---}

\texttt{<input>}

\end{tcolorbox}

\begin{tcolorbox}[
    breakable,
    title=Prompt Template for Event Suggestions Generation,
    colback=white,
    colframe=black!75,
    boxrule=0.5pt,
    left=4pt, right=4pt, top=2pt, bottom=2pt,
    arc=2mm,
    fonttitle=\bfseries,
]
\paragraph{1. Role and Goal}
You are an expert public opinion analyst acting as a strategic consultant. Your task is to generate a set of practical, actionable suggestions based on a thorough analysis of the provided report sections: the \textit{Title}, \textit{Summary}, \textit{Timeline}, and \textit{Focus}.

\paragraph{2. Reasoning Process and Content Requirements}
To formulate your suggestions, you must follow this reasoning process:
\begin{enumerate}
    \item \textbf{Synthesize the Context:} First, thoroughly analyze all provided report sections to build a comprehensive understanding of the event's theme, its evolution, and the different viewpoints of Netizens and Authoritative Institutions.
    \item \textbf{Identify Core Issues:} Based on your analysis, pinpoint the key problems, public concerns, and negative sentiments that need to be addressed.
    \item \textbf{Generate Actionable Suggestions:} Formulate a set of clear and concise suggestions (Max 200 words total) that directly address the identified issues. Your recommendations should be practical and aim to manage the situation, improve public perception, or enhance engagement.
\end{enumerate}

\paragraph{3. Few-Shot Examples}
\begin{itemize}
    \item \textbf{Input Report Sections:} \verb|{example_report_sections_input}|
    \item \textbf{Output Suggestions:} \verb|{example_suggestions_output}|
\end{itemize}

\hrulefill

\begin{center}\textbf{[PROVIDED REPORT SECTIONS]}\end{center}

\textbf{Event\_Title:}
\texttt{<title\_text>}

\vspace{1em}
\textbf{Event\_Summary:}
\texttt{<summary\_text>}

\vspace{1em}
\textbf{Event\_Timeline:}
\texttt{<timeline\_text>}

\vspace{1em}
\textbf{Event\_Focus:}
\texttt{<focus\_text>}

\end{tcolorbox}

\subsection{Prompts for Evaluation Framework (\evaluationname)} \label{app:prompts-evaluation}
The following prompts and guidelines are used by the \evaluationname~ framework to systematically evaluate the generated reports. The framework consists of a main agent that manages three specialist tools.
\subsubsection{Main Prompt for the \evaluationname Agent} \label{app:prompt-agent}
\begin{tcolorbox}[
    breakable,
    title=Prompt for the Evaluation Agent,
    colback=white,
    colframe=black!75,
    boxrule=0.5pt,
    left=4pt, right=4pt, top=2pt, bottom=2pt,
    arc=2mm,
    fonttitle=\bfseries,
]


Try your best to evaluate the quality of the given public opinion report comprehensively.

<tool introduction>

You have access to three specialized evaluation tools:

\textbf{Fact-Checker Tool:} Verifies factual accuracy by comparing report content against reference data ($Z_i$). Use this for Event Title, Event Summary, and Timeline date accuracy.

\textbf{Opinion Mining Tool:} Analyzes public opinion coverage by examining social media posts ($Y_i$). Use this for Timeline coverage completeness and Event Focus evaluation.

\textbf{Solution Counselor Tool:} Evaluates recommendation quality using your expert knowledge. Use this for Event Suggestions evaluation.

\textbf{Use the following format:}

\textbf{Initial Input:} the public opinion report to be evaluated. If the report is too long, focus on the section relevant to current evaluation.

\textbf{Thought:} analyze which aspect needs to be evaluated and which tool to use.

\textbf{Tool to Use:} should be one of [Fact-Checker, Opinion-Mining, Solution-Counselor]

\textbf{Tool Input:} the specific content for the selected tool

\textbf{Observation:} the evaluation score (1-5) with reasoning from the tool

... (this Thought/Tool to Use/Tool Input/Observation can repeat N times for each report section)

\textbf{Thought:} I have completed evaluating all sections and can provide final scores

\textbf{Final Scores:} The final output for the i-th report is a 15-dimensional score vector $S_i = (s_{i,1}, \dots, s_{i,15})$ where:

\begin{itemize}
    \setlength\itemsep{-0.5em}
    \item $s_{i,1}$: Title score (from Fact-Checker Tool)
    \item $s_{i,2}$ to $s_{i,6}$: Summary scores (from Fact-Checker Tool)
    \item $s_{i,7}$: Timeline Date Accuracy score (from Fact-Checker Tool)
    \item $s_{i,8}$: Timeline Coverage score (from Opinion Mining Tool)
    \item $s_{i,9}$ to $s_{i,11}$: Focus scores (from Opinion Mining Tool)
    \item $s_{i,12}$ to $s_{i,15}$: Suggestion scores (from Solution Counselor Tool)
\end{itemize}

<in-context examples>

Begin!

<report to evaluate>

\end{tcolorbox}

\section{Experiment}\label{app:experiment}

\subsection{Two-Phase Human Evaluation Procedure} \label{app:human_eval_phases}
Our human evaluation protocol is executed in two distinct phases to ensure the highest quality and reliability of the ratings.

\paragraph{Phase 1: Annotator Calibration}
To ensure all three experts share a consistent and unified understanding of criteria, we use a preliminary set of 50 reports (5 events by our five baseline models under both generation strategies). The three experts independently rate this calibration set, and we calculate the Intraclass Correlation Coefficient (ICC). 

\paragraph{Phase 2: Formal Evaluation}
The formal evaluation phase then uses a distinct and larger corpus of 500 reports (from 50 events). 
The three calibrated experts then independently score this entire corpus, yielding three complete sets of ratings for our human-agent agreement analysis.

\subsection{Human Evaluation Tool Interfaces}

\begin{figure}[h]
    \centering
    \includegraphics[width=0.5\textwidth]{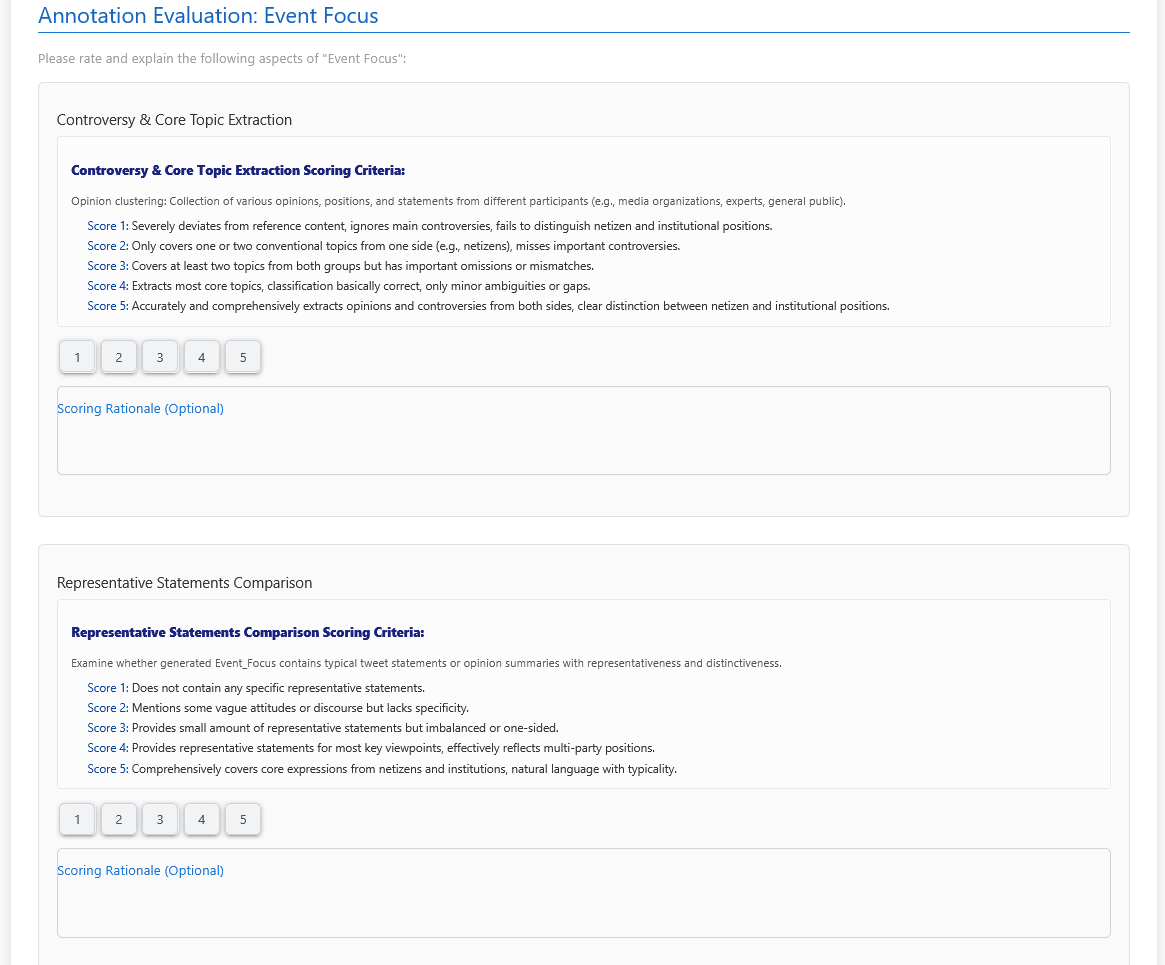}
    \caption{Screenshot of the human evaluation tool interface for the Event Focus.}
    \label{fig:human_evaluation_tool}
\end{figure}

\subsection{Main Results}\label{app:main-results}

\clearpage 

\begin{landscape}
\thispagestyle{empty} 
\begin{table}[H]
    \centering
    \small
    \setlength{\tabcolsep}{4pt}
    \begin{tabular}{ll@{\hspace{8pt}}l|r|rrrrr|rr|rrr|rrrr}

        \toprule
        & & & & \multicolumn{5}{c|}{\textbf{Event Summary}} & \multicolumn{2}{c|}{\textbf{Event Timeline}} & \multicolumn{3}{c|}{\textbf{Event Focus}} & \multicolumn{4}{c}{\textbf{Event Suggestions}} \\
        
        \cmidrule{5-9} \cmidrule{10-11} \cmidrule{12-14} \cmidrule{15-18}
        
        \multirow{-2}{*}{\textbf{Evaluator}} & \multirow{-2}{*}{\textbf{\shortstack{Gen\\Strategy}}} & \multirow{-2}{*}{\textbf{\parbox{1.5cm}{\centering Model}}} & \multirow{-2}{*}{\textbf{\shortstack{Event\\Title}}} & \textbf{\shortstack{Event\\Nature}} & \textbf{\shortstack{Time \&\\Loc.}} & \textbf{\shortstack{Involved\\Parties}} & \textbf{Causes} & \textbf{Impact} & \textbf{\shortstack{Date\\Acc.}} & \textbf{\shortstack{Sub\\Events}} & \textbf{\shortstack{Contro.\\Topic}} & \textbf{\shortstack{Repr.\\Stmt.}} & \textbf{\shortstack{Emo.\\Anal.}} & \textbf{Rel.} & \textbf{Feas.} & \textbf{\shortstack{Emo.\\Guide.}} & \textbf{Innov.} \\

        \midrule
        \multirow{10}{*}{DeepSeek-V3} 
         & \multirow{5}{*}{End-to-end} 
         & DeepSeek-R1 & \textbf{4.48} & \textbf{4.77} & 4.52 & \textbf{3.55} & \textbf{4.26} & 4.15 & 2.68 & 3.73 & 4.23 & 4.06 & 3.91 & 4.95 & \textbf{4.00} & 3.73 & \textbf{3.98} \\
         & & DeepSeek-V3 & 4.37 & 4.72 & 4.54 & 3.36 & 4.12 & 4.21 & 2.14 & \textbf{3.85} & 4.13 & 4.04 & 3.88 & 4.93 & 3.99 & 3.60 & 3.96 \\
         & & Gemini 2.5 Pro & 4.37 & 4.76 & \textbf{4.61} & 3.46 & 4.22 & 4.30 & \textbf{3.31} & 3.48 & \textbf{4.42} & \textbf{4.42} & \textbf{3.96} & \textbf{4.97} & \textbf{4.00} & \textbf{3.85} & \textbf{3.98} \\
         & & GPT-4o & 4.29 & 4.60 & 4.45 & 3.34 & 3.93 & 4.16 & 2.48 & 3.63 & 4.18 & 4.02 & 3.88 & 4.80 & 3.99 & 3.49 & 3.92 \\
         & & Llama-3.3-70B & 4.15 & 4.57 & 4.35 & 3.12 & 3.39 & \textbf{4.31} & 2.00 & 3.34 & 3.98 & 3.80 & 3.88 & 4.65 & 3.97 & 3.28 & 3.89 \\
         \cmidrule{2-18}
         & \multirow{5}{*}{Modular} 
         & DeepSeek-R1 & 4.34 & \textbf{4.70} & 4.34 & \textbf{3.57} & \textbf{4.17} & 4.12 & \textbf{3.60} & 3.51 & 4.42 & \textbf{4.39} & 3.93 & \textbf{4.98} & \textbf{4.00} & \textbf{3.96} & \textbf{3.94} \\
         & & DeepSeek-V3 & 4.28 & 4.67 & 4.35 & 3.37 & \textbf{4.17} & 4.15 & 3.47 & 3.45 & 4.57 & 4.20 & 3.94 & 4.94 & 3.98 & 3.91 & 3.85 \\
         & & Gemini 2.5 Pro & \textbf{4.38} & \textbf{4.70} & \textbf{4.42} & 3.36 & 4.12 & \textbf{4.16} & 3.57 & 3.47 & \textbf{4.63} & 4.28 & \textbf{3.95} & 4.94 & \textbf{4.00} & 3.86 & 3.89 \\
         & & GPT-4o & 4.12 & 4.52 & 4.22 & 3.23 & 3.84 & 4.12 & 3.24 & 3.57 & 4.39 & 4.29 & 3.94 & 4.84 & \textbf{4.00} & 3.77 & 3.87 \\
         & & Llama-3.3-70B & 4.15 & 4.40 & 4.11 & 3.14 & 3.68 & 3.98 & 2.80 & \textbf{3.60} & 4.37 & 4.27 & 3.92 & 4.76 & 3.99 & 3.51 & 3.83 \\
         \midrule
        \multirow{10}{*}{GPT-4o} 
         & \multirow{5}{*}{End-to-end} 
         & DeepSeek-R1 & \textbf{4.50} & 4.41 & 4.02 & \textbf{2.78} & 3.68 & 3.36 & 1.22 & 3.89 & 3.81 & 3.66 & 3.79 & 4.67 & 3.97 & 3.57 & 3.74 \\
         & & DeepSeek-V3 & 4.40 & 4.36 & 4.00 & 2.48 & 3.49 & 3.55 & 1.25 & 3.78 & 3.81 & 3.64 & 3.80 & \textbf{4.80} & 3.98 & 3.58 & 3.58 \\
         & & Gemini 2.5 Pro & 4.47 & \textbf{4.46} & \textbf{4.12} & 2.77 & \textbf{3.76} & 3.52 & 1.27 & \textbf{4.15} & \textbf{3.85} & \textbf{3.88} & 3.83 & 4.79 & \textbf{3.99} & \textbf{3.71} & 3.53 \\
         & & GPT-4o & 4.33 & 4.28 & 3.97 & 2.51 & 3.34 & 3.51 & \textbf{1.34} & 3.78 & 3.81 & 3.67 & \textbf{3.84} & 4.57 & 3.98 & 3.51 & \textbf{3.77} \\
         & & Llama-3.3-70B & 4.27 & 4.22 & 3.89 & 2.38 & 2.93 & \textbf{3.67} & 1.14 & 3.33 & 3.76 & 3.48 & 3.78 & 4.67 & 3.95 & 3.56 & 3.51 \\
         \cmidrule{2-18}
         & \multirow{5}{*}{Modular} 
         & DeepSeek-R1 & 4.16 & 4.28 & 3.81 & \textbf{2.72} & \textbf{3.56} & 3.37 & 1.27 & 4.08 & \textbf{3.88} & 3.82 & 3.83 & 4.62 & 3.99 & \textbf{3.98} & 3.43 \\
         & & DeepSeek-V3 & 4.14 & 4.28 & 3.82 & 2.53 & 3.55 & 3.53 & 1.20 & 4.07 & 3.82 & 3.76 & 3.77 & 4.57 & 3.96 & 3.97 & 3.37 \\
         & & Gemini 2.5 Pro & \textbf{4.26} & \textbf{4.34} & \textbf{3.93} & 2.56 & 3.53 & 3.49 & \textbf{1.28} & \textbf{4.12} & 3.85 & \textbf{3.84} & 3.84 & \textbf{4.70} & \textbf{4.00} & 3.92 & 3.30 \\
         & & GPT-4o & 4.04 & 4.08 & 3.71 & 2.50 & 3.24 & \textbf{3.55} & 1.27 & 4.10 & \textbf{3.88} & 3.78 & \textbf{3.85} & 4.40 & 3.98 & 3.89 & 3.62 \\
         & & Llama-3.3-70B & 4.13 & 3.97 & 3.59 & 2.36 & 3.06 & 3.37 & 1.24 & 3.89 & 3.78 & 3.69 & 3.77 & 4.45 & 3.95 & 3.73 & \textbf{3.63} \\
        \bottomrule
    \end{tabular}
    \caption{\textbf{Detailed} performance comparison across all 15 evaluation sub-dimensions for five LLMs using two generation strategies (end-to-end and modular), evaluated by two distinct LLM evaluators (DeepSeek-V3 and GPT-4o). This table expands the five main dimensions from Table~\ref{tab:performance_summary_0} into their constituent sub-dimensions. Within each experimental block, the highest score for each sub-dimension is highlighted in \textbf{bold}.}
    \label{tab:performance_summary_1}
\end{table}
\end{landscape}

\clearpage 


\end{document}